\documentclass[sn-mathphys-num]{sn-jnl}% Math and Physical Sciences Numbered Reference Style 
\usepackage{graphicx}%
\usepackage{multirow}%
\usepackage{amsmath,amssymb,amsfonts}%
\usepackage{amsthm}%
\usepackage{mathrsfs}%
\usepackage[title]{appendix}%
\usepackage{xcolor}%
\usepackage{textcomp}%
\usepackage{manyfoot}%
\usepackage{booktabs}%
\usepackage{algorithm}%
\usepackage{algorithmicx}%
\usepackage{algpseudocode}%
\usepackage{listings}%
\usepackage{subfig}
\usepackage{float}
\usepackage{dutchcal}
\DeclareMathOperator{\st}{s.t.}
\theoremstyle{thmstyleone}%
\theoremstyle{thmstyletwo}%

\theoremstyle{thmstylethree}%

\begin{document}

\title[Article Title]{Point Neuron Learning: A New Physics-Informed Neural Network Architecture}

%%=============================================================%%
%% GivenName	-> \fnm{Joergen W.}
%% Particle	-> \spfx{van der} -> surname prefix
%% FamilyName	-> \sur{Ploeg}
%% Suffix	-> \sfx{IV}
%% \author*[1,2]{\fnm{Joergen W.} \spfx{van der} \sur{Ploeg} 
%%  \sfx{IV}}\email{iauthor@gmail.com}
%%=============================================================%%

\author*[1]{\fnm{Hanwen} \sur{Bi}}\email{hanwen.bi@anu.edu.au}

\author[1]{\fnm{Thushara} \sur{D. Abhayapala}}\email{thushara.abhayapala@anu.edu.au}
%\equalcont{These authors contributed equally to this work.}

%\author[1,2]{\fnm{Third} \sur{Author}}\email{iiiauthor@gmail.com}
%\equalcont{These authors contributed equally to this work.}

\affil[1]{\orgdiv{Audio \& Acoustics Signal Processing Group, School of Engineering}, \orgname{The Australian National University}, \orgaddress{\city{Canberra}, \postcode{2600}, \state{ACT}, \country{Australia}}}

%\affil[2]{\orgdiv{Department}, \orgname{Organization}, \orgaddress{\street{Street}, \city{City}, \postcode{10587}, \state{State}, \country{Country}}}

%\affil[3]{\orgdiv{Department}, \orgname{Organization}, \orgaddress{\street{Street}, \city{City}, \postcode{610101}, \state{State}, \country{Country}}}

%%==================================%%
%% Sample for unstructured abstract %%
%%==================================%%

\abstract{Machine learning and neural networks have advanced numerous research domains, but challenges such as large training data requirements and inconsistent model performance hinder their application in certain scientific problems.
To overcome these challenges, researchers have investigated integrating physics principles into machine learning models, mainly through: (i) physics-guided loss functions, generally termed as physics-informed neural networks, and (ii) physics-guided architectural design. 
While both approaches have demonstrated success across multiple scientific disciplines, they have limitations including being trapped to a local minimum, poor interpretability, and restricted generalizability.
This paper proposes a new physics-informed neural network (PINN) architecture that combines the strengths of both approaches by embedding the fundamental solution of the wave equation into the network architecture, enabling the learned model to strictly satisfy the wave equation. 
The proposed point neuron learning method can model an arbitrary sound field based on microphone observations without any dataset. 
Compared to other PINN methods, our approach directly processes complex numbers and offers better interpretability and generalizability.
We evaluate the versatility of the proposed architecture by a sound field reconstruction problem in a reverberant environment.
Results indicate that the point neuron method outperforms two competing methods and can efficiently handle noisy environments with sparse microphone observations.}


%%\pacs[JEL Classification]{D8, H51}

%%\pacs[MSC Classification]{35A01, 65L10, 65L12, 65L20, 65L70}

\maketitle

\section{Introduction}\label{sec1}
With the explosive growth of available data and computing resources, machine learning and neural networks have been greatly developed and successfully implemented in many research disciplines \cite{he2016deep,voulodimos2018deep,goldberg2022neural,sherstinsky2020fundamentals,dwivedi2024octant, dwivedi2023long,luo2023delayless,lecun2015deep,ribeiro2023sound}. However, in some scientific domains, including acoustics, a large number of observation data is infeasible, and the learned model is unable to produce physically consistent results. To address these challenges, researchers have explored integrating scientific principles into machine learning models, typically through two main approaches: physics-guided loss functions and physics-guided architectural design. Although both approaches 
%physics-guided loss functions and physics-guided architectural design 
have been successfully applied to various discipline specific problems, they have some limitations including being trapped to a local minimum or converged to a trivial solution \cite{feng2023physics, wong2022learning}, poor interpretability \cite{willard2020integrating}, and limited ability to generalize to out-of-sample scenarios \cite{willard2020integrating}. In this paper, we aim to leverage the advantages of both approaches by proposing a neural network architecture that ensures strict adherence to the governing equations from physics, thereby enhancing interpretability and generalizability.

The idea of physics-guided loss functions was proposed in physics-informed neural networks (PINNs) \cite{raissi2017physics,raissi2019physics}, where partial differential equations (PDEs) related loss terms are added to the neural networks. In the typical PINN, other than the observed data, more data are created to calculate the PDE loss, which can enhance the model robustness and help the learned model fit in the governed PDE. PINN has been widely implemented in diverse scientific fields, including fluid physics \cite{cai2021flow,li2020coupled,erichson2019physics,raissi2020hidden,tartakovsky2020physics,zhu2021general,rasht2022physics}, thermodynamics \cite{patel2022thermodynamically,jia2021physics,kim2022physics}, and medical imaging \cite{sarabian2022physics,kissas2020machine,hammernik2023physics,van2022physics}.
%and is becoming one of the most popular research topics. 
%Physics-informed neural networks (PINNs) \cite{raissi2017physics,raissi2019physics} was proposed to overcome the small data plight for problems governed by partial differential equations (PDEs). In the typical PINN, other than the observed data, more data are created to calculate the PDE loss, which can enhance the model robustness and help the learned model fit in the governed PDE. PINN has been widely implemented in diverse scientific fields, including fluid physics \cite{cai2021flow,li2020coupled,erichson2019physics,raissi2020hidden,tartakovsky2020physics,zhu2021general,rasht2022physics}, thermodynamics \cite{patel2022thermodynamically,jia2021physics,kim2022physics}, and medical imaging \cite{sarabian2022physics,kissas2020machine,hammernik2023physics,van2022physics}, and is becoming one of the most popular research topics. 

Physics-guided architectural design involves integrating principles from physics into neural network architectures to enhance model performance, data efficiency, and interpretability. One straightforward approach is to embed conventional methods into the processing layers of neural networks \cite{robinson2022physics}. Amos and Kolter added a differentiable convex optimization solver layer into the proposed OptNet \cite{amos2017optnet}. In the proposed differentiable physics engine, a rigid body simulator was embedded into the network architecture to enable the network to learn physical parameters from data \cite{de2018end}. Another strategy is encoding invariances and symmetries into network architecture \cite{willard2020integrating}. Ling \textit{et al} demonstrated that the neural network with the rotational invariance property can improve the prediction accuracy of turbulence modelling \cite{ling2016reynolds}. Anderson \textit{et al} designed a rotationally covariant neural network architecture for better learning of the behaviour and properties of complex many-body physical systems \cite{anderson2019cormorant}.

In the field of acoustics, PINNs have demonstrated successful applications, while research on physics-guided architectural design remains relatively unexplored. Shigemi \textit{et al}. integrated PINN with a bicubic spline interpolation for the sound field estimation problem \cite{shigemi2022physics}. Borrel-Jensen, Engsig-Karup, and Jeong implemented PINN with different boundary conditions to predict one-dimensional sound fields \cite{borrel2021physics}. For the problem of nearfield acoustics holography, Olivieri \textit{et al} proposed a Kirchhoff–Helmholtz-based convolutional neural network (KHCNN) which involved Kirchhoff–Helmholtz equation to calculate loss functions \cite{olivieri2021physics}, and Kafri \textit{et al} further combined KHCNN with the famous explainable CNN architecture Gradient-weighted Class Activation Mapping (Grad-CAM) \cite{selvaraju2017grad} to make it more explainable \cite{kafri2023grad}. For the room impulse response reconstruction, Pezzoli, Antonacci, and Sarti used PINN to reconstruct the early part of room impulse response in time-domain \cite{pezzoli2023implicit}, and the network in \cite{karakonstantis2024room,karakonstantis2023room} is able to estimate particle velocity and intensity, in addition to sound pressure. PINNs have also contributed to head-related transfer functions \cite{ma2023physics}, sound field estimation or reconstructions \cite{chen2023sound,ma2023circumvent,karakonstantis2023advancing,olivieri2024physics}, active noise control \cite{10447208}, and ocean acoustics \cite{yoon2024predicting,yoon2023physics}.

Although PINNs have demonstrated great potential in the field of acoustics, the idea of adding the PDE loss to existing neural network architectures has some limitations. First, for the frequency-domain approach, most neural networks cannot directly process complex numbers, and the separated training of real and imaginary numbers has the risk of missing phase information. Second, even though the wave equation is involved in the training process, the learned model is still a black box model and hardly interpretable. Third, the working range of the learned model is restricted by the microphone observations or training dataset, resulting in the difficulty to generalize the learned model to out-of-sample scenarios \cite{willard2020integrating}, especially for sound field extrapolation problems \cite{williams1999fourier,bi2023spherical}.

As the PINN is a generalized approach for all PDE-determined systems, the direct adoption of PINN cannot fully utilize features behind the wave equation, and one outstanding feature of the wave equation is that its fundamental solution is known. Therefore, rather than only adding PDE losses in neural networks, we embed the fundamental solution, free space Green function \cite{skudrzyk2012foundations}, into the network architecture, enabling the learned model to strictly satisfy the wave equation. In the proposed network, the basic processing unit is called a {\em point neuron} whose weight and biases can be learned by back propagation. The physical meaning of point neurons is equivalent to point sources or plane wave sources, and the weight and distribution of equivalent sources can be updated while training. The proposed {\em point neuron network} can be implemented to model and estimate an arbitrary sound field purely based on microphone observations without an extra dataset. %, no extra dataset is required.

The main contribution of this paper is four-fold. First, the proposed method provides a new way to integrate principles from physics with neural networks. Second, the proposed method can directly process complex numbers, which can maintain the phase information of complex sound fields. Third, the point neuron learning is interpretable in physics, and the nature of equivalent sources enables the generalizability to out-of-sample scenarios. Finally, the training process only requires a small number of microphone observations, and no extra dataset or pre-training process is required. 

The rest of this paper is organized as follows. In Section \ref{sec2}, we formulate a generalized sound field modelling problem. We present the point neuron learning architecture and derive the back propagation for it in Section \ref{sec3}.  In  Section \ref{sec4}, we evaluate the proposed method with a sound field estimation problem in a reverberant environment and compare its performance with one conventional method and one typical PINN method. Finally, Section \ref{sec5} concludes this paper.

%%%%%%%%%%%%%%%%%%%%%%%%%%%%%%%%%%%%%%%%%%%%%%%%%%%%%%%%%%%%%%%%%%%%%%%
%%%%%%%%%%%%  Sec2: Problem Formulation
%%%%%%%%%%%%%%%%%%%%%%%%%%%%%%%%%%%%%%%%%%%%%%%%%%%%%%%%%%%%%%%%%%%%%%%%

\section{Problem formulation}\label{sec2}
\begin{figure}[!t]
\centering
\includegraphics[width=3.4in]{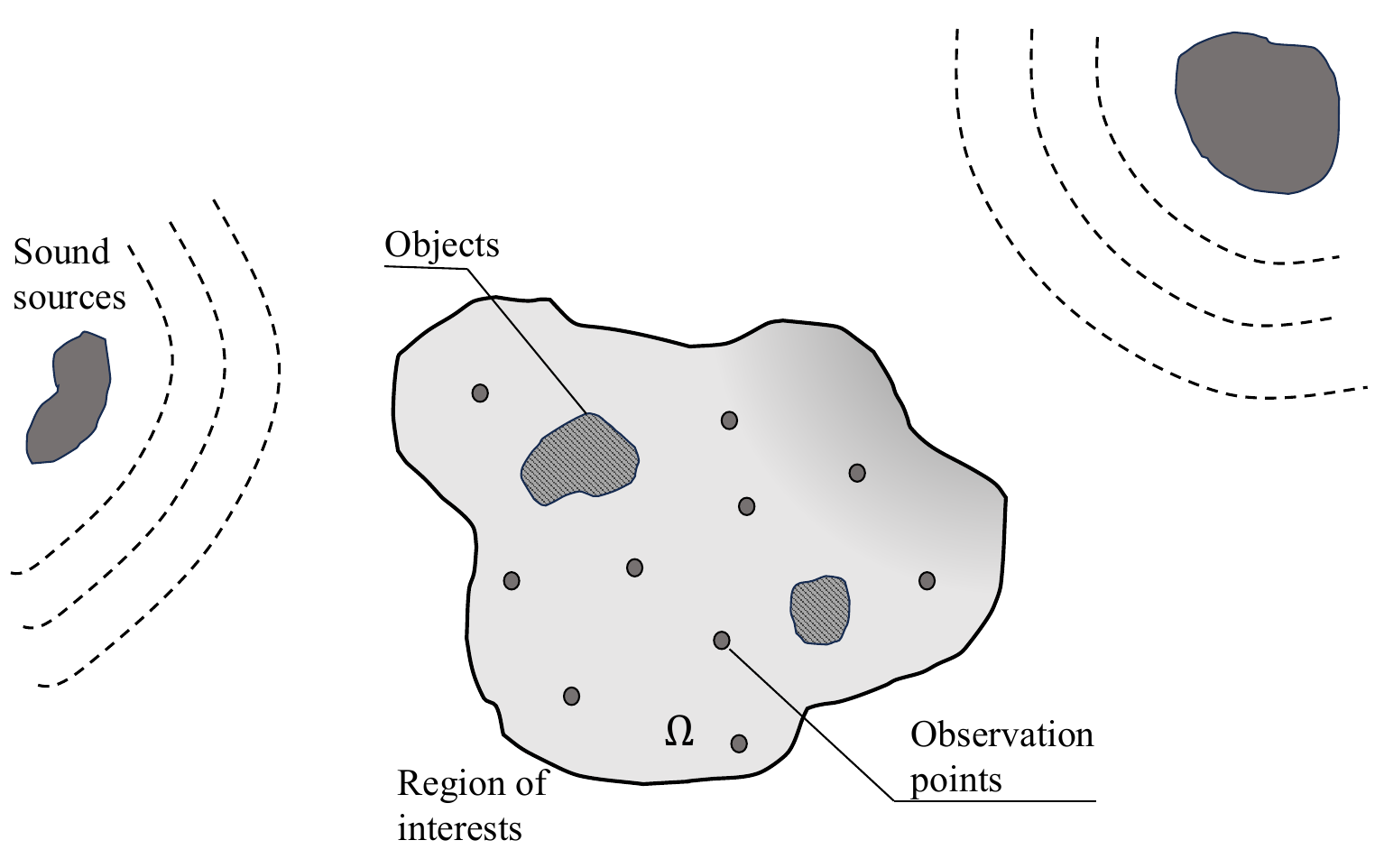}
\caption{Illustration of the target region $\Omega$, observation points, and sound sources.}
\label{fig_1}
\end{figure}

Consider a source-free region of space denoted by $\Omega \subseteq \mathbb{R}^{2 \mathrm{or} 3}$ surrounded by sound sources as shown in Fig.~\ref{fig_1}. 
The sound pressure at an arbitrary point in the Cartesian coordinate $\boldsymbol x=(x, y, z), \boldsymbol x \in \Omega$, can be represented by $P(\boldsymbol x, k)\in \mathbb{C}$, where $k=2\pi f/c$ is the wave number, $f$ %\in \mathbb{R}$ 
is the frequency, and $c$ is the speed of sound propagation.
Let there are $Q$ number of observation points $\boldsymbol x_{q} \in \Omega$, $q = 1,...,Q$, to measure the sound field,  as shown in Fig.~\ref{fig_1}.  

We aim to build a neural network model $\mathcal{P}(\boldsymbol x, k; \boldsymbol \mu)$ for the sound field $P(\boldsymbol x, k)$ over the target region $\Omega$ from $Q$ observation points, where $\boldsymbol\mu \in\mathbb{C}^{S}$ are the parameters of the model, $S$ is the number of model parameters. The model $\mathcal{P}(\boldsymbol x, k; \boldsymbol \mu)$ should satisfy the Helmholtz equation which governs the wave propagation over space. We formulate $\mathcal{P}(\boldsymbol x, k; \boldsymbol \mu)$ by the following optimization problem, 
\begin{subequations}\label{eq:1}
\label{formulation}
\begin{align}
&\underset{\boldsymbol\mu\in\mathbb{C}^{S}}{\arg\min} && \mathcal{L}=\mathcal{L}_{\text{TRN}}(\boldsymbol{p}_{\mathrm{mic}},\mathbcal{p}_{\mathrm{mic}})+\lambda \mathcal{C}(\boldsymbol \mu), \label{eq:1A}\\
&\st &&\Delta^{2}\mathcal{P}(\boldsymbol x, k; \boldsymbol \mu)+k^{2}\mathcal{P}(\boldsymbol x, k; \boldsymbol \mu)=0 \label{eq:1B}\\
&&&\boldsymbol x \in \Omega \nonumber\\
&&&\lambda \in [0,\infty) \nonumber
\end{align}
\end{subequations}
where $\mathcal{L}$ is the cost function, $\mathcal{L}_{\text{TRN}}$ is the training loss that measures the supervised error between the sound field estimation 
%$\mathbcal{p}_{mic}=[\mathcal{P}(\boldsymbol x_{1}, k), \mathcal{P}(\boldsymbol x_{2}, k), ... , \mathcal{P}(\boldsymbol x_{q}, k),...,\mathcal{P}(\boldsymbol x_{Q}, k)]^T$ 
$\mathbcal{p}_{\mathrm{mic}}=[\mathcal{P}(\boldsymbol x_{1}, k), \ldots, \mathcal{P}(\boldsymbol x_{Q}, k)]^T$ 
and the pressure measurements 
%$\boldsymbol{p}_{q}=[P(\boldsymbol x_{1}, k), P(\boldsymbol x_{2}, k), ... , P(\boldsymbol x_{q}, k), ... ,P(\boldsymbol x_{Q}, k)]^T$,
$\boldsymbol{p}_{\mathrm{mic}}=[P(\boldsymbol x_{1}, k), \ldots, P(\boldsymbol x_{Q}, k)]^T$,  $\lambda$ is a hyper-parameter to  control the weight of model complexity loss $\mathcal{C}(\boldsymbol \mu)$,
and
%%%%%
\[
%$
\Delta^2\equiv \frac{\partial^{2}}{\partial x^{2}}+\frac{\partial^{2}}{\partial y^{2}}+\frac{\partial^{2}}{\partial z^{2}}.
%$
\]
%%%%
Then, the research problem becomes how to find the optimal model $\mathcal{P}(\boldsymbol x, k; \boldsymbol \mu)$ based on $Q$ observations that can minimize $\mathcal{L}$ as well as satisfy the wave propagation constrain in (\ref{eq:1B}).
%To evaluate the model, $M$ number of evaluation points are placed in $\Omega$, $m=1,...,M$ be the index of evaluation points, and $\boldsymbol x_{m} \in \Omega$. The model prediction error $\epsilon(k)$ is defined by 

%$\epsilon(k)$ can be used to assess the sound field model. 

%%%%%%%%%%%%%%%%%%%%%%%%%%%%%%%%%%%%%%%%%%%%%%%%%%%%%%%%%%%%%%%%%%%%%%%
%%%%%%%%%%%%  Sec3: Point Neuron Learning
%%%%%%%%%%%%%%%%%%%%%%%%%%%%%%%%%%%%%%%%%%%%%%%%%%%%%%%%%%%%%%%%%%%%%%%%
\section{Point neuron learning}\label{sec3}
In this section, we propose a new network architecture, termed the {\em point neuron network}, which satisfies the Helmholtz equation constraint, and the learned model is explainable using principles in physics.
%
%%%%%%%%%%%%%%%%%%%%
\subsection{Point neuron}
%%%%%%%%%%%%%%%%%%%%%%%%%
%
The propagation of sound is governed by the Helmholtz wave equation (\ref{eq:1B}) which is a natural constraint of any valid sound field and limits the direct adaptation of existing neural network architectures to model the sound field. Therefore, to satisfy (\ref{eq:1B}), we embed the fundamental solution of the Helmholtz equation, i.e., the free space Green function,  into a neural network architecture, 
%
%%%%%%%%%%%%%%
\begin{equation}
\label{Green}
G(\boldsymbol x|\boldsymbol y, k)=\frac{e^{ik\rVert\boldsymbol x-\boldsymbol y\rVert_{2}}}{4\pi\rVert\boldsymbol x-\boldsymbol y\rVert_{2}},
\end{equation}
%%%%%%%%%%%%%
%
where $i=\sqrt{-1}$, $\rVert\cdot\rVert_{2}$ is $\ell_2$-norm, $\boldsymbol y$ denotes the location of an omni-directional unit point source, $G(\boldsymbol x|\boldsymbol y, k)$ means the sound field at an observer point $\boldsymbol x$ generated by a unit strength point source at $\boldsymbol y$.

We add a normalization to unify the treatment of both near-field and far-field sources and make it the building block of our network, named {\em point neuron unit}, expressed as
\begin{equation}
\label{point neuron}
PN(\boldsymbol x|\boldsymbol y, k)=\rVert \boldsymbol y\rVert_{2} e^{-ik\rVert \boldsymbol y\rVert_{2}}\frac{e^{ik\rVert\boldsymbol x-\boldsymbol y\rVert_{2}}}{4\pi\rVert\boldsymbol x-\boldsymbol y\rVert_{2}}.
\end{equation}
The point neuron unit works as a virtual source that is capable of modelling both near-field and far-field sound propagation \cite{birnie2021mixed}. 
\begin{figure}[!t]
\centering
\includegraphics[width=3.4in]{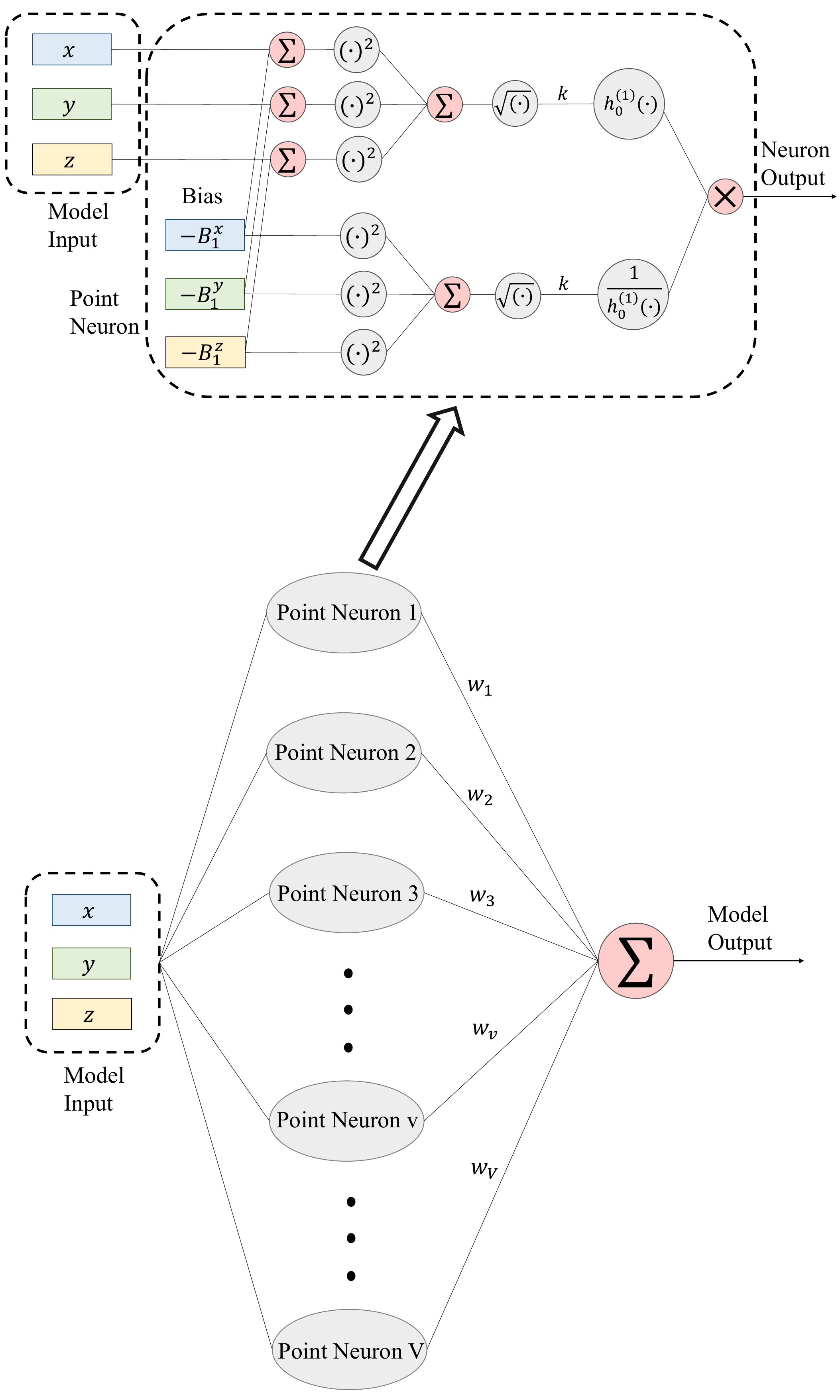}
\caption{Network architecture of point neuron learning. Top: the point neuron building block. Bottom: the architecture overview with $V$ number of point neurons.}
\label{fig_2}
\end{figure}
%
%%%%%%%%%%%%%%%%%%%%
%\subsection{Point Neuron network}
%%%%%%%%%%%%%%%%%%%%%%%%%
%
As shown in Fig.~\ref{fig_2},  we embed V number of point neuron units into a neural networks architecture. The network input is the coordinate of an arbitrary point $\boldsymbol x=(x, y, z)$, and the output is the sound pressure of that point $P(\boldsymbol x, k)$. 

The $v^{\text{th}}$  point neuron unit has three inputs, $x$, $y$, and $z$ co-ordinates of the point $\boldsymbol{x}$, and three biases $B^{x}_{v}, B^{y}_{v}, \mathrm{and} ~B^{z}_{v} \in \mathbb{R}$. 
%are added to each channel of input. 
In the top path, the three channels of inputs with biases first go through an activation function $(\cdot)^{2}$ individually and then are fully connected (with unity weights) to pass another activation function $\sqrt{(\cdot)}$ and multiplying with the wave number $k$ (can be considered as a constant weight) before passing 
%The fixed weight $k$ is added before it is passed 
through the final activation function $h^{(1)}_{0}(\cdot)$, %where $h_{0}(\cdot)$ 
which is the $0^{\text{th}}$ order spherical Hankel function of the first kind, given by 
\begin{equation}
\label{Hankel}
h^{(1)}_{0}(r)=\frac{e^{ir}}{ir}.
\end{equation}
%%%%%%%%%%%
%
In the bottom path, the three biases directly undergo the same process as in the top path, followed by the final activation function $1/h^{(1)}_{0}(\cdot)$ after multiplying the fixed weight $k$. Finally, the outputs of the top and bottom paths are multiplied together to generate the point neuron output. We construct the point neuron network by fully connecting $V$ point neuron units with neuron weights $w_{v}\in \mathbb{C}$, $v=1,\ldots, V$ as shown in Fig.~\ref{fig_2}. The point neuron and its weights are frequency-dependent.

%Finally, the outputs of each point neuron unit are fully connected with the neuron weight $w_{v}\in \mathbb{C}$ to perform the model output.

Note that the physical meaning of biases and the weight of the point neuron unit is the location and strength of the corresponding virtual source. The weight and distribution of virtual sources can be learned by the training process. Though the point neuron architecture is a direct mapping from the normalized Green function, the resulting network can be learned (i.e., weights and biases) as any other neural network using established tools.

%
%%%%%%%%%%%%%%%%%%%%%%%%%%%%%%%%%%%%%%%%
\subsection{Back propagation and training}
%%%%%%%%%%%%%%%%%%%%%%%%%%%%%%%%%%%%%
%
According to the network architecture and (\ref{eq:1}), we define the system cost function as 
\begin{subequations}\label{eq:2}
\label{formulation}
\begin{align}
&\underset{\boldsymbol\mu\in\mathbb{C}^{S}}{\arg\min} && \mathcal{L}=\sum^{Q}_{q=1}{| \mathcal{P}(\boldsymbol x_{q}, k; \boldsymbol \mu)-P(\boldsymbol x_{q}, k) |^{2}}+\lambda\rVert \boldsymbol w \rVert_{1}, \label{eq:2A}\\
&\st &&\Delta^{2}\mathcal{P}(\boldsymbol x, k; \boldsymbol \mu)+k^{2}\mathcal{P}(\boldsymbol x, k; \boldsymbol \mu)=0 \label{eq:2B}\\
&&&\boldsymbol x \in \Omega \nonumber\\
&&&\lambda \in [0,\infty) \nonumber
\end{align}
\end{subequations}
where  $\rVert\cdot\rVert_{1}$ is $\ell_1$-norm,  $\boldsymbol w= [w_{1}, \ldots, w_{V}]^{T}$ is a $V \times 1$ vector that contains the neuron weights. We apply $\ell_1$-norm as the model complexity loss to control the number of activated point neurons and to avoid overfitting. Note that all network parameters are frequency/wave number dependent and for convenience, we omit the dependency on $k$  for the rest of the paper.

The network parameters can be updated iteratively by back propagation. In specific, for the $v^{\text{th}}$ point neuron, weights $w_{v}$ can be updated by
%
%%%%%%%%%%%%%%
\begin{equation}
\label{update w}
w_{v}(n+1)=w_{v}(n)-\xi\frac{\partial\mathcal{L}(n)}{\partial w_{v}(n)^{*}},
\end{equation}
%%%%%%%%%%%%%
%
where 
\begin{equation}
\begin{aligned}
\label{gradient w}
\frac{\partial\mathcal{L}(n)}{\partial w_{v}^{*}(n)}=&\sum_{q=1}^{Q}\Big(\mathcal{P}(\boldsymbol x_{q}, n)-P(\boldsymbol x_{q})\Big)\,\frac{ D_{v}(n)}{D_{v}^{q}(n)} \,e^{-ik\big(D_{v}^{q}(n)-D_{v}(n)\big)}\,
+\frac{1}{2}\lambda e^{i\theta_{v}(n)},
\end{aligned}
\end{equation}
with
\begin{align}
D_{v}(n) &=\sqrt{{B^{x}_{v}(n)}^{2}+{B^{y}_{v}(n)}^{2}+{B^{z}_{v}(n)}^{2}},\nonumber \\
D^{v}_{q}(n) &=\sqrt{({B^{x}_{v}(n)-x_q)}^{2}+{(B^{y}_{v}(n)-y_q)}^{2}+{(B^{z}_{v}(n)-z_{q})}^{2}},\nonumber
\end{align}
% \begin{equation}
% D^{v}_{q}(n)=\sqrt{({B^{x}_{v}(n)-x_q)}^{2}+{(B^{y}_{v}(n)-y_q)}^{2}+{(B^{z}_{v}(n)-z_{q})}^{2}},\nonumber
% \end{equation}
$n$ is the iteration index of the training process, $\xi$ is the learning rate, $(\cdot)^{*}$ denotes complex conjugation, and $\theta_{v}(n)\in \mathbb{R}$ is the phases of complex weight $w_{v}$. The proof of \eqref{gradient w} is given in the Appendix~\ref{secA1}.

Using back propagation, we can also update bias parameters $B_{v}^{\alpha}(\cdot)$ where $\alpha \in \{x,y,z\}$  by 
\begin{equation}
\label{update B}
%B_{v}^{x,y,\mathrm{or}~z}(n+1)=B_{v}^{x,y,\mathrm{or}~z}(n)-\xi\frac{\partial\mathcal{L}(n)}{\partial B_{v}^{x,y,\mathrm{or}~z}(n)},
B_{v}^{\alpha}(n+1)=B_{v}^{\alpha}(n)-\xi\frac{\partial\mathcal{L}(n)}{\partial B_{v}^{\alpha}(n)},
\end{equation}
%where $\alpha \in \{x,y,z\}$, and
where
\begin{equation}
\begin{small}
\begin{aligned}
\label{gradient Bx}
\frac{\partial\mathcal{L}(n)}{\partial B_{v}^{x}(n)}=&\sum_{q=1}^{Q}~2\operatorname{\mathbb{R}e}\Bigg\{\big(\mathcal{P}(\boldsymbol x_{q}, n)-P(\boldsymbol x_{q})\big)^{*}~w_{v}~\frac{ D_{v}(n)}{D_{v}^{q}(n)}~e^{ik\big(D_{v}^{q}(n)-D_{v}(n)\big)}~\\
\times&\Big(\frac{-\big(ikD_{v}(n)-1\big)}{{D_{v}(n)}^2}B_{v}^{x}(n)+\frac{\big(ikD_{v}^{q}(n)-1\big)}{{D_{v}^{q}(n)}^{2}}\big(B_{v}^{x}(n)-x_{q}\big)\Big)\Bigg\},
\end{aligned}
\end{small}
\end{equation}
%%%%%%%%%%%%%%%%%%%%%%%%%%%%%%%%%%%%%%%%%%%%%%%%%%%
\begin{equation}
\begin{small}
\begin{aligned}
\label{gradient By}
%\frac{\partial\mathcal{L}(n)}{\partial B_{v}^{y}(n)}=&\sum_{q=1}^{Q}~2\mathcal{R}~
%\Bigg(\frac{w_{v}(n)e^{ik\big(D_{v}^{q}(n)-D_{v}(n)\big)}}{D_{v}^{q}(n)}\Big(\frac{-\big(ikD_{v}(n)-1\big)}{D_{v}(n)}B_{v}^{y}(n)\\
%+&\frac{\big(ikD_{v}^{q}(n)-1\big)D_{v}(n)}{{D_{v}^{q}(n)}^{2}}\big(B_{v}^{y}(n)-y_{q}\big)\Big)\big(\mathcal{P}(\boldsymbol x_{q}, n)-P(\boldsymbol x_{q})\big)^{*}\Bigg),
\frac{\partial\mathcal{L}(n)}{\partial B_{v}^{y}(n)}=&\sum_{q=1}^{Q} 2\,  \operatorname{\mathbb{R}e}\Bigg\{\big(\mathcal{P}(\boldsymbol x_{q}, n)-P(\boldsymbol x_{q})\big)^{*}~w_{v}~\frac{ D_{v}(n)}{D_{v}^{q}(n)}~e^{ik\big(D_{v}^{q}(n)-D_{v}(n)\big)}~\\
\times&\Big(\frac{-\big(ikD_{v}(n)-1\big)}{{D_{v}(n)}^2}B_{v}^{y}(n)+\frac{\big(ikD_{v}^{q}(n)-1\big)}{{D_{v}^{q}(n)}^{2}}\big(B_{v}^{y}(n)-y_{q}\big)\Big)\Bigg\},
\end{aligned}
\end{small}
\end{equation}
%%%%%%%%%%%%%%%%%%%%%%%%%%%%%%%%%%%%%%%%%%%%%%%%%%%
\begin{equation}
\begin{small}
\begin{aligned}
\label{gradient Bz}
\frac{\partial\mathcal{L}(n)}{\partial B_{v}^{z}(n)}=&\sum_{q=1}^{Q}~2\operatorname{\mathbb{R}e}\Bigg\{\big(\mathcal{P}(\boldsymbol x_{q}, n)-P(\boldsymbol x_{q})\big)^{*}~w_{v}~\frac{ D_{v}(n)}{D_{v}^{q}(n)}~e^{ik\big(D_{v}^{q}(n)-D_{v}(n)\big)}~\\
\times&\Big(\frac{-\big(ikD_{v}(n)-1\big)}{{D_{v}(n)}^2}B_{v}^{z}(n)+\frac{\big(ikD_{v}^{q}(n)-1\big)}{{D_{v}^{q}(n)}^{2}}\big(B_{v}^{z}(n)-z_{q}\big)\Big)\Bigg\},
\end{aligned}
\end{small}
\end{equation}
where $\operatorname{\mathbb{R}e}\{ \cdot\}$ represents the real part of the argument. The proofs of \eqref{gradient Bx}, \eqref{gradient By} and \eqref{gradient Bz} are given in Appendix~\ref{secA2}.
%By (\ref{update w}) to (\ref{gradient Bz}), system parameters can be updated iteratively.

We have the following comments:
\begin{enumerate}
    \item In the training process, only microphone observations are used to learn the model, and no extra data or datasets are required. Therefore, the proposed network architecture is feasible in the small data regime where datasets are not accessible, which is common in real sound field model scenarios. 
    \item From (\ref{gradient w}), (\ref{gradient Bx}), (\ref{gradient By}), and (\ref{gradient Bz}), gradients of $w_{v}(n)$, $B_{v}^{x}(n)$, $B_{v}^{y}(n)$, and $B_{v}^{z}(n)$ become infinity when $D_{v}(n)$ and $D_{v}^{q}(n)$ becomes zero, which means virtual sources cannot be located at observer point positions or the origin point. Considering the physical meaning of the network, observer points cannot be on the source positions. Thus, during the training process, we apply a strategy to avoid placing virtual sources on the observer positions nor at the origin.  We calculate the $D_{v}(n)$ and $D_{v}^{q}(n)$ in each iteration and replace $B_{v}^{x}(n)$, $B_{v}^{y}(n)$, and $B_{v}^{z}(n)$ to another location when $D_{v}(n)$ or $D_{v}^{q}(n)$ is smaller than a certain threshold. 
\end{enumerate}

%%%%%%%%%%%
\subsection{System initialization}
%%%%%%%%%%%%%%
We initialise point neuron weights $w_v$ such that $|w_v|$ in the range of $[-1,1]$. 
%We apply different strategies to initialize the point neuron weights and system biases.   For the initialization of point neuron weights $w_v$
For the initialization of system biases, 
we use any Prior-known knowledge about the underlying physical scenario to determine the initial positions of the virtual sources. For example, for modelling the outgoing sound field of a certain sound source, such as a drone, the virtual sources can be initially placed within a sphere that encompasses the sound source. The initialization will be further discussed in the next section.

%%%%%%%%%%%%%%%%%%%%%%%%%%%%%%%%%%%%%%%%%%%%%%%%%%%%%%%%%%%%%%%%%%%%%%%
%%%%%%%%%%%%  Sec4: Experiments
%%%%%%%%%%%%%%%%%%%%%%%%%%%%%%%%%%%%%%%%%%%%%%%%%%%%%%%%%%%%%%%%%%%%%%%%
\section{Simulation experiments}\label{sec4}
%%%%%%%%%%%%%%%%%%%%%%%%%%%%%%
%
\begin{figure*}[!ht]
\centering
\subfloat[]{\includegraphics[width=0.5\textwidth]{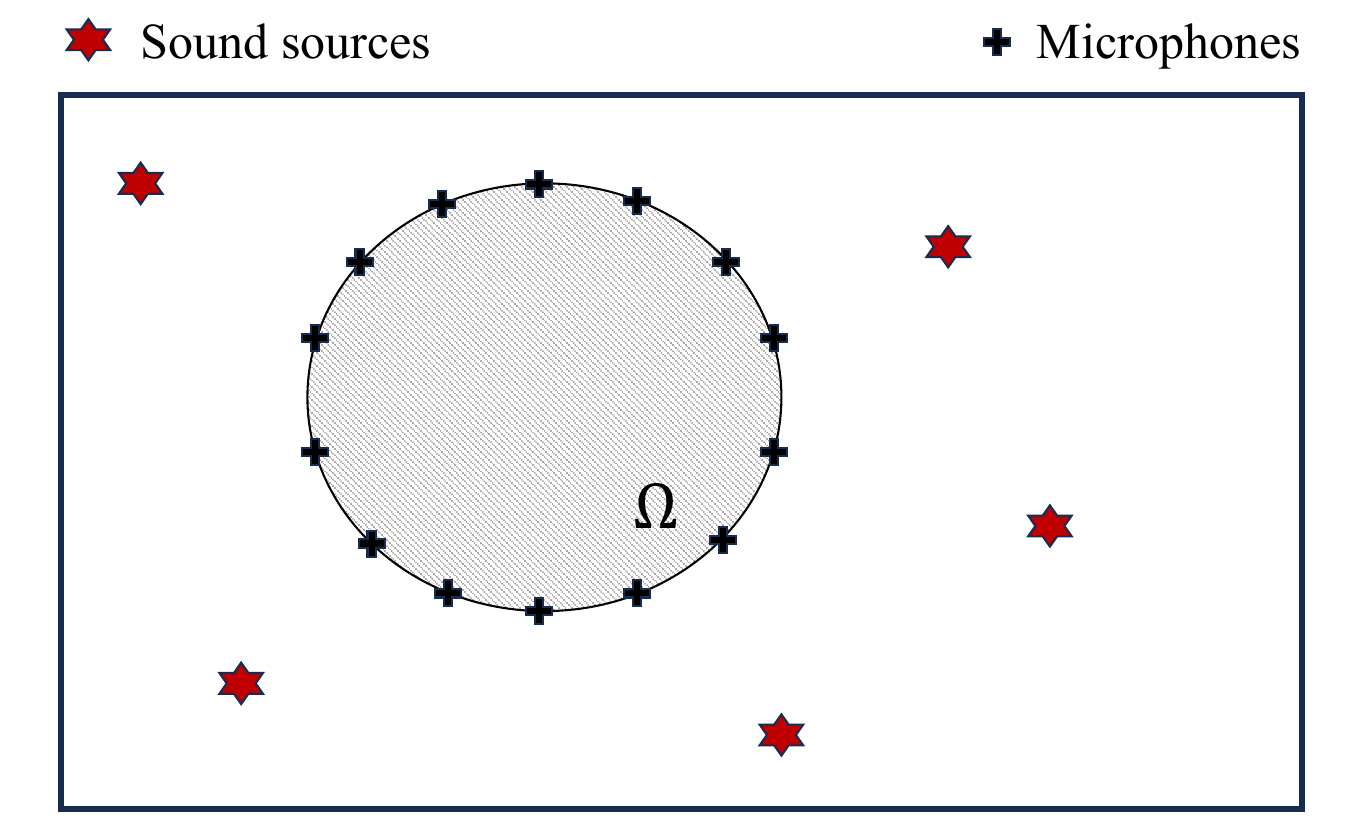}%
\label{figure3a}}
\hfill
\subfloat[]{\includegraphics[width=0.5\textwidth]{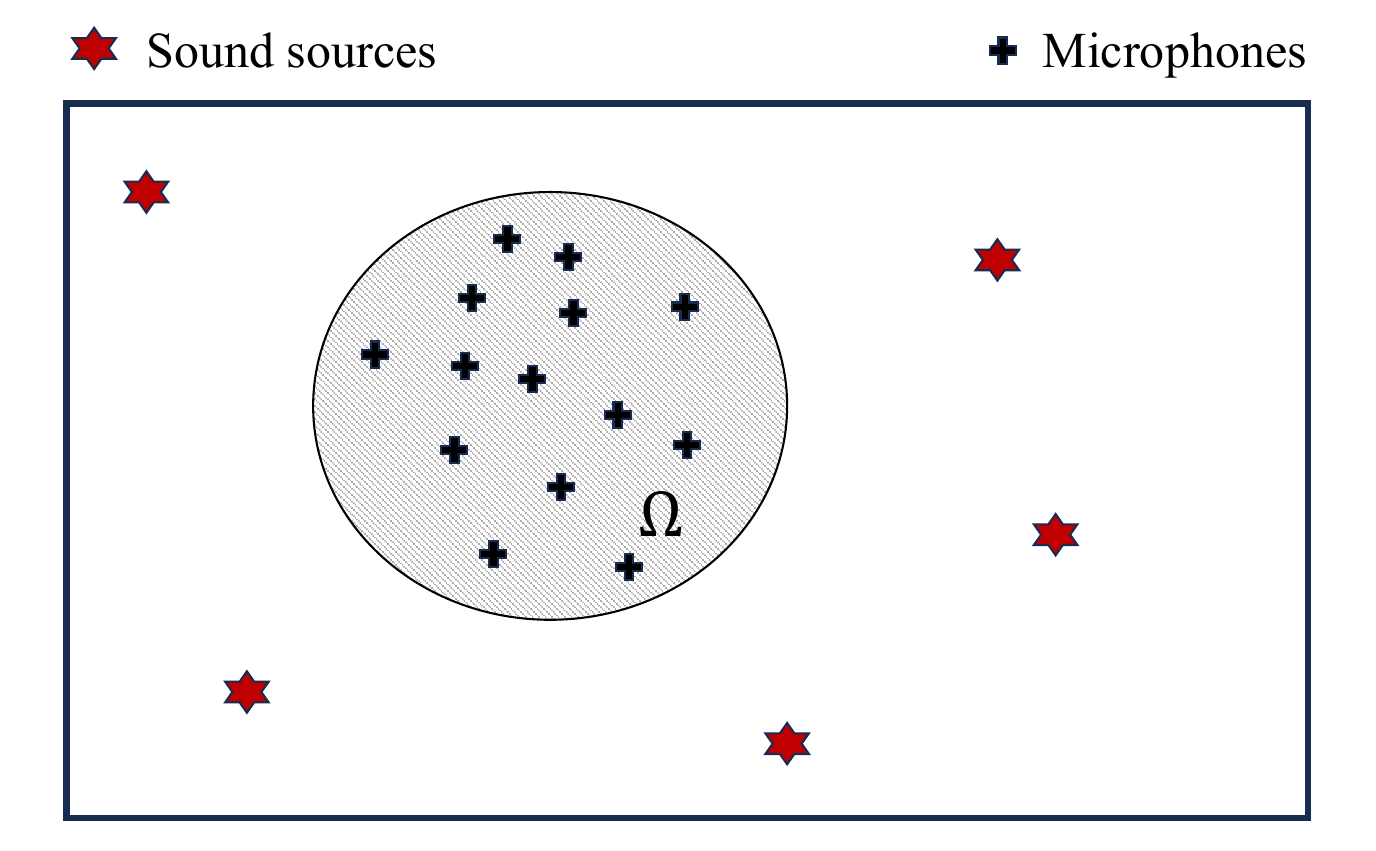}%
\label{figure3b}}
\caption{Experimental setup. The sound field over the circular target region $\Omega$ is estimated. (a) Microphones are uniformly placed over the edge of the target region. (b) Microphones are arbitrarily placed in the target region.}
\label{simulation setup}
\end{figure*}
To illustrate the versatility of the proposed point neuron network architecture, we apply it to the sound field reconstruction problem where the goal is to accurately estimate the sound field over a spatial region using a limited set of measurements from microphones.  

\subsection{Room settings}
We evaluate the proposed method in a rectangular reverberant room with dimensions $6.0$ m $\times$ $4.0$ m $\times$ $4.0$ m where the coordinate origin is in the middle of the room. The target reconstruction region $\Omega$ is a horizontal circular region with a radius of $R=1.0$ m, centred at the point $(-1.0, 0.5, 0.0)$ m.  There are five sources with equal strength (set to unity) located at the same horizontal plane, with coordinates $(-2.65, 1.5, 0.0)$ m, $(-2.4, -1.2, 0.0)$ m, $(0.2, -1.5, 0.0)$ m, $(1.7, -0.2, 0.0)$ m, and $(1.0, 1.2, 0.0)$ m. We use the image source method \cite{allen1979image} with wall reflection coefficients $[0.8, 0.8, 0.8, 0.8, 0, 0]$ to simulate the sound field over the target reconstruction region. The corresponding reverberation time $RT_{60}$ is $0.537$ s \cite{habets2006room}.  Note that, we set the reflection coefficients of the roof and ground to zero to simplify the illustration.

% We evaluate the proposed method in a reverberant environment. The setup is shown in Figure~\ref{simulation setup}. The room is a rectangular room with a size of 6.0 m $\times$ 4.0 m $\times$ 4.0 m, centered at the origin. The target region $\Omega$ is a horizontal circular region with a radius of $R=1.0$ m, centered at the point (2.0, 2.5, 2.0) m. Five unit strength point sources are located at the same horizontal plane, with coordinates (0.35, 3.5, 2.0) m, (0.6, 0.8, 2.0) m, (3.2, 0.5, 2.0) m, (4.7, 1.8, 2.0) m, and (4.0, 3.2, 2.0) m. The sound field over the reverberant room is simulated by the image source method \cite{allen1979image}. We choose the wall reflection coefficients as $[0.8, 0.8, 0.8, 0.8, 0, 0]$, with 0 reflection coefficients for the roof and ground, and the related reverberation time $RT_{60}$ is 0.537 s \cite{habets2006room}. 

We consider two different microphone placements to capture the sound field: (i) uniform circular array on the boundary of the target region (Fig.~\ref{figure3a}), and (ii) randomly selected locations inside the target region (Fig.~\ref{figure3b}). We initially place  $Q=75$ microphones based on $Q=2N+1$ where $N=\lceil{k_{f_{\text{max}}}R}\rceil$ and $\lceil{\cdot}\rceil$ is the ceiling operation \cite{kennedy2007intrinsic}.  In Sec.~\ref{sec:mic_num}, we investigate the performance by varying the number of microphones. For all simulations, except Sec.~\ref{sec:different_snr}, White Gaussian noise with an SNR of 20 dB is added to each of the microphone measurements.

For reconstruction performance evaluations, we consider the frequency band from $100$ to $2000$ Hz with $100$ Hz increments and use a grid of uniformly spaced 1124 evaluation points over the target region, with point separation of $5.3$ cm.

%a grid size of 0.053 m. 

\subsection{Network settings}

We randomly initialize the network weights by $| w_{v}|\in [-1,1]$. For the system biases, we initially place them over a mesh grid on the same horizontal plane as $\Omega$, excluding the $\Omega$, with a size of 9.0 m $\times$ 9.0 m. The number of point neurons increases from $V=25$ for the lowest frequency to $V=465$ for the highest frequency. The parameter $\lambda$ is in the range of $[3\times 10^{-4}, 3\times 10^{-3}]$, and $\xi=3\times 10^{-2}$. 

% A total 1124 evaluation points are uniformly sampled in the target region, with a grid size of 0.053 m. Two different microphone placements are investigated in this section. For the first kind of placement, as shown in Figure~\ref{figure3a}, $Q$ number of microphones are uniformly placed over the boundary of the target region. For the second kind of placement, as shown in Figure~\ref{figure3b}, $Q$ number of microphones are randomly placed inside the target region. The frequency band from 100 to 2000 Hz with 100 Hz increments are evaluated. We initially select the microphone number $Q=75$ based on $Q=2N+1$ and $N=\lceil{k_{f_{max}}R}\rceil$ ($\lceil{\cdot}\rceil$ is the ceiling operation) \cite{kennedy2007intrinsic}. White Gaussian noise with an SNR of 20 dB is added to each of the microphone measurements. We randomly initialize the network weights by $\rVert w_{v}\rVert_{2}\in [-1,1]$. For the system biases, we initially place them over a mesh grid on the same horizontal plane as $\Omega$, excluding the $\Omega$, with a size of 9.0 m $\times$ 9.0 m. The point neuron number $V$ increases from $V=25$ for the lowest frequency to $V=465$ for the highest frequency. The parameter $\lambda$ is in the range of $[3\times 10^{-4}, 3\times 10^{-3}]$, and $\xi=3\times 10^{-2}$. 

We compare the proposed sound field reconstruction with a conventional method and a typical PINN method. The conventional method is the orthogonal harmonics-based sound field estimation method that reconstructs any incident sound field by a finite number of harmonic basis functions and corresponding coefficients \cite{wu2008theory,williams1999fourier}. We adopt the network architecture proposed in \cite{ma2023circumvent} as the typical PINN method since it adds PDE loss to a fully connected network architecture similar to many PINN methods. Similar to the proposed method, the PINN in \cite{ma2023circumvent} also only requires microphone observations in the training process. This network contains 3 layers and 15 nodes of each layer, with the activation function being $tanh$, and initializes the trainable parameters with the Xavier initialization. We train the PINN for $108$ epochs with a learning rate of $10^{-5}$ using the ADAM optimizer. The data loss of the network is calculated by all microphones, and the PDE loss is calculated by $437$ uniformly arranged sampling points over the target region. 

% We compare the proposed method with a conventional method and a typical PINN method. The conventional method is the harmonics-based sound field estimation method that reconstructs any incident sound field by a finite number of harmonic modes and corresponding coefficients \cite{wu2008theory,williams1999fourier}. For the typical PINN method, we adopt a similar network architect proposed in \cite{ma2023circumvent}, because it is a typical implementation of PINN that adds PDE loss to a fully connected network architecture. Similar to the proposed method, the PINN in \cite{ma2023circumvent} also only requires microphone observations in the training process. The network contains 3 layers and 15 nodes of each layer, with the activation function being $tanh$, and initialize the trainable parameters with the Xavier initialization. We train the PINN for 108 epochs with a learning rate of $10^{-5}$ using the ADAM optimizer. The data loss of the network is calculated by all microphones, and the PDE loss is calculated by 437 uniformly arranged sampling points over the target region. 

\subsection{Evaluation metrics}
Two different metrics are used to evaluate the overall model estimation. The first metric is the normalized mean square error (NMSE) between the original sound field $P(\boldsymbol x, k)$ and the sound field model $\mathcal{P}(\boldsymbol x, k)$ for each frequency point $k$, which is defined by
\begin{equation}
\label{NMSE}
\mathrm{NMSE}(k)=20\mathrm{log}_{10}\frac{\sum^{M}_{m=1}|\mathcal{P}(\boldsymbol x_{m}, k)-P(\boldsymbol x_{m}, k)|}{\sum^{M}_{m=1}| P(\boldsymbol x_{m}, k)|},
\end{equation}
where $M$ is the number of evaluation points, $m=1,...,M$ be the index of evaluation points, and $\boldsymbol x_{m} \in \Omega$.

The second metric is the Modal Assurance Criterion (MAC) \cite{pastor2012modal} which can assess the overall similarity between the estimated sound field and the original sound field for each frequency. The MAC value is a real scalar bounded by 0 and 1. When the estimated sound field shapes correspond to the original sound field the resultant MAC estimate is close to 1, but it never reaches 1 unless the same mode shape vectors are compared. $\mathrm{MAC}(k)$ can be calculated by
\begin{equation}
\label{MAC}
\mathrm{MAC}(k)=\frac{\rVert\boldsymbol{p}_{\mathrm{eval}}^{H}\mathbcal{p}_{\mathrm{eval}}\rVert_{2}^{2}}{(\boldsymbol{p}_{\mathrm{eval}}^{H}\boldsymbol{p}_{\mathrm{eval}})(\mathbcal{p}_{\mathrm{eval}}^{H}\mathbcal{p}_{\mathrm{eval}})}, 
\end{equation}
where $\boldsymbol{p}_{\mathrm{eval}}=[P(\boldsymbol x_{1}, k), \ldots, P(\boldsymbol x_{M}, k)]^T$ and $\mathbcal{p}_{\mathrm{eval}}=[\mathcal{P}(\boldsymbol x_{1}, k),\ldots,\mathcal{P}(\boldsymbol x_{M}, k)]^T$ are two $M \times 1$ vectors that contain the original sound pressure and estimated sound pressure of evaluate points, $(\cdot)^H$ denotes the conjugate transpose.

We also visualize the error distribution of the estimated sound field for certain frequencies. The corresponding normalized square error (NSE) is defined as
\begin{equation}
\label{NSE}
\mathrm{NSE}(\boldsymbol x,k)=20\mathrm{log}_{10}\frac{|\mathcal{P}(\boldsymbol x, k)-P(\boldsymbol x, k) |}{| P(\boldsymbol x, k)|}.
\end{equation}

%\subsection{Results with the circular microphone placement}\label{subsection4-2}
\subsection{Circular microphone array placement}\label{subsection4-2}
\begin{figure*}[!ht]
\centering
\subfloat[]{\includegraphics[width=0.48\textwidth]{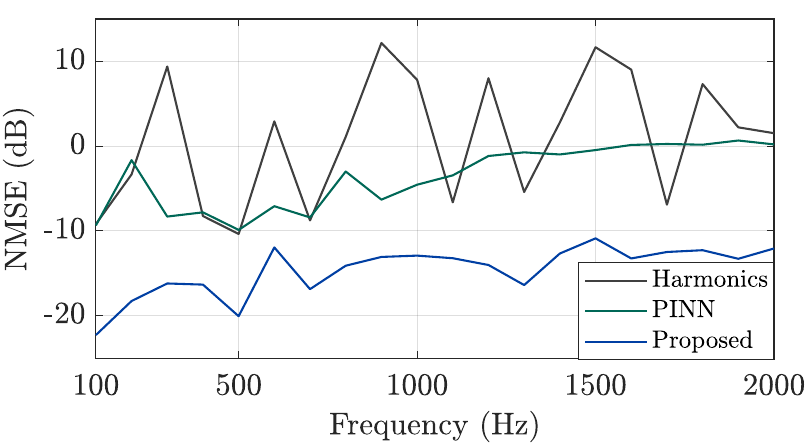}%
\label{figure4a}}
\hfill
\subfloat[]{\includegraphics[width=0.48\textwidth]{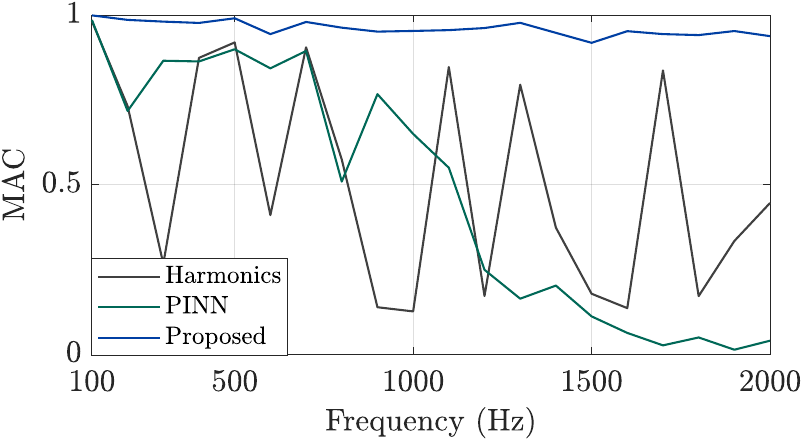}%
\label{figure4b}}
\caption{NMSE and MAC with respect to the frequency with circular microphone placement: (a) NMSE (b) MAC.}
\label{ErrorPlotCir}
\end{figure*}

%%%%%%%%%%%%%%%%%%%
%
Figure~\ref{ErrorPlotCir} depicts the performance metrics NMSE \eqref{NMSE} and MAC \eqref{MAC} of the proposed method, harmonics-based method, and PINN method as a function of frequency.  Overall, the proposed method outperforms the other two methods for all frequencies, especially when the frequency is higher than 1100 Hz. Specifically, based on the NMSE result as shown in Fig.~\ref{figure4a}, (i) the proposed method has the lowest NMSE for all frequencies, and it slightly increases from $-22$ dB in $100$ Hz to $-13$ dB in $2000$ Hz, (ii) 
the PINN method can roughly estimate the sound field under $1100$ Hz, while above $1100$ Hz the NMSE reaches $0$ dB, and (iii) the NMSE for the harmonics-based method fluctuates between $-10$ dB to $10$ dB, and peaks are caused by the zeros of the spherical Bessel function. For small target regions, the harmonics-based method is reliable, but for larger target regions, more basis functions are used to estimate the sound field, and more Bessel zeros are involved in the estimation, resulting in poor performance of the harmonic-based method.

% The NMSE and MAC assessment results for the proposed method, harmonics-based method, and PINN method are shown in Figure~\ref{ErrorPlotCir}. Overall, the proposed method outperforms competing methods for all frequencies, especially when the frequency is higher than 1100 Hz. In specific, based on the NMSE result as shown in Figure~\ref{figure4a}, i) the proposed method has the lowest NMSE for all frequencies, and it slightly increases from -22 dB in 100 Hz to -13 dB in 2000 Hz; ii) The PINN method can roughly estimate sound field under 1100 Hz, while above 1100 Hz the NMSE is reaching 0 dB; iii) the NMSE for harmonics-based method bounces between -10 dB to 10 dB, and peaks are caused by the zeros of Bessel function. For small target regions, the harmonics-based method is reliable, but for larger target regions, more Bessel modes are used to estimate the sound field, and more Bessel zeros are involved in the estimation, resulting in the poor performance of the harmonic-based method in this estimation process.

From Fig.~\ref{figure4b},  we find that (i) the proposed method has the highest MAC over all frequencies; (ii) the performance of PINN method degrades after $1100$ Hz, indicating that the PINN model cannot estimate the sound field distribution for higher frequencies; (iii) the MAC results for harmonics-based method fluctuates over a large range, indicating the harmonics based method can only work for certain frequencies with the given microphone placement.
%(iv) compared with Fig.~\ref{figure4a}, for higher frequencies, the NMSE result for the harmonic-based method fluctuates around the PINN method, while the MAC result for the harmonics-based method totally outperforms the PINN method. % (I am not sure about this observation)

Since the dimensionality of the sound field increases with frequency, we observe that the given number of microphones may be not enough to train the PINN to learn the sound field, resulting in its poor performance at higher frequencies. Moreover, the low MAC value for the PINN method over high frequencies also reveals the potential of losing phase information for the learned model.

% The metric MAC against the frequency is plotted in Fig.~\ref{figure4b}, where we find that (i) the proposed method has the highest MAC over all frequencies; (ii) the performance of PINN method collapses after $1100$ Hz, indicating the PINN model cannot estimate the sound field distribution with higher frequencies; (iii) the MAC results for harmonics-based method fluctuates in a large range, indicating the harmonics based method can only work for certain frequencies with the given microphone placement; (iv) compared with Fig.~\ref{figure4a}, for higher frequencies, the NMSE result for the harmonic-based method fluctuates around the PINN method, while the MAC result for the harmonics-based method totally outperforms the PINN method. For higher frequencies, the the sound field is more complicated, and the given number of microphones may be not enough to train the PINN to learn the sound field, resulting in its poor performance in higher frequencies. Moreover, the low MAC value for the PINN method over high frequencies also reveals the potential of losing phase information for the learned model.

\begin{figure*}[!t]
\centering
\subfloat[]{\includegraphics[width=0.25\textwidth]{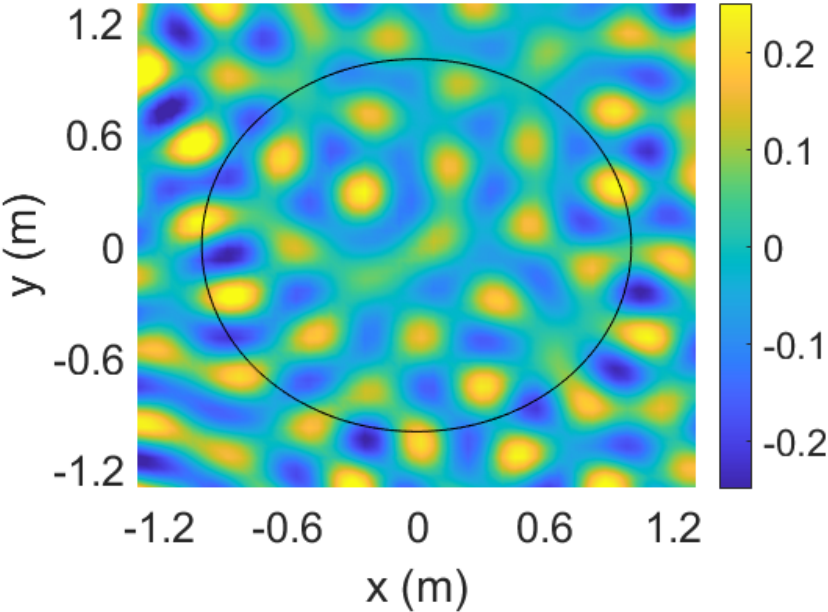}%
\label{figure5a}}
\subfloat[]{\includegraphics[width=0.25\textwidth]{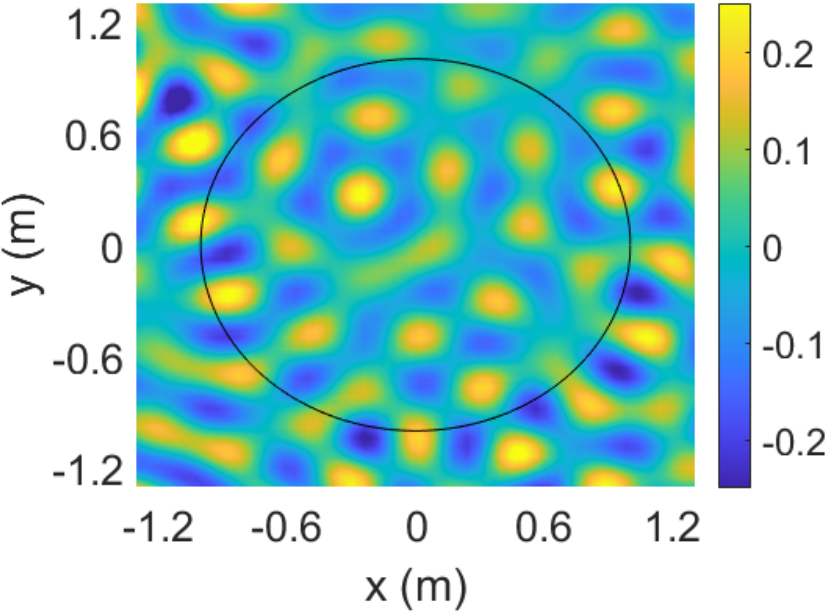}%
\label{figure5b}}
\subfloat[]{\includegraphics[width=0.25\textwidth]{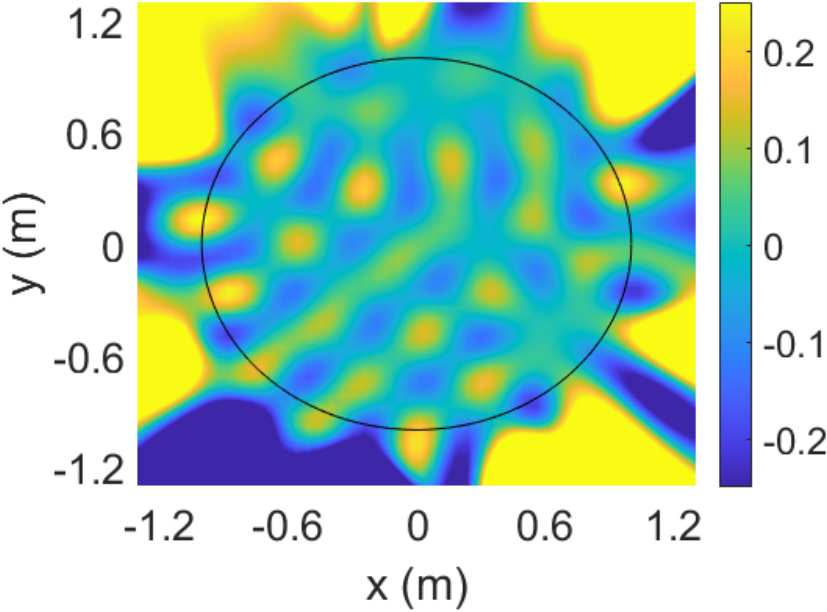}%
\label{figure5c}}
\subfloat[]{\includegraphics[width=0.25\textwidth]{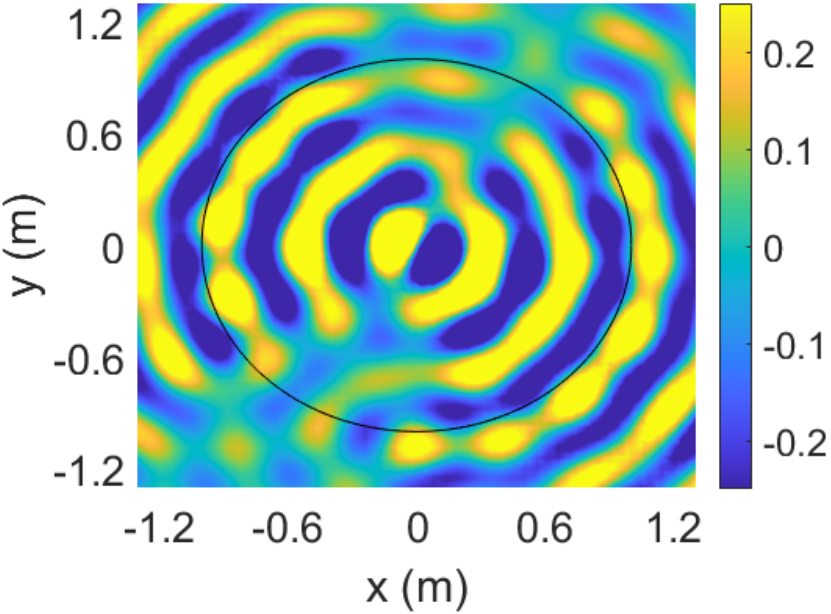}%
\label{figure5d}}
\hfil
\subfloat[]{\includegraphics[width=0.25\textwidth]{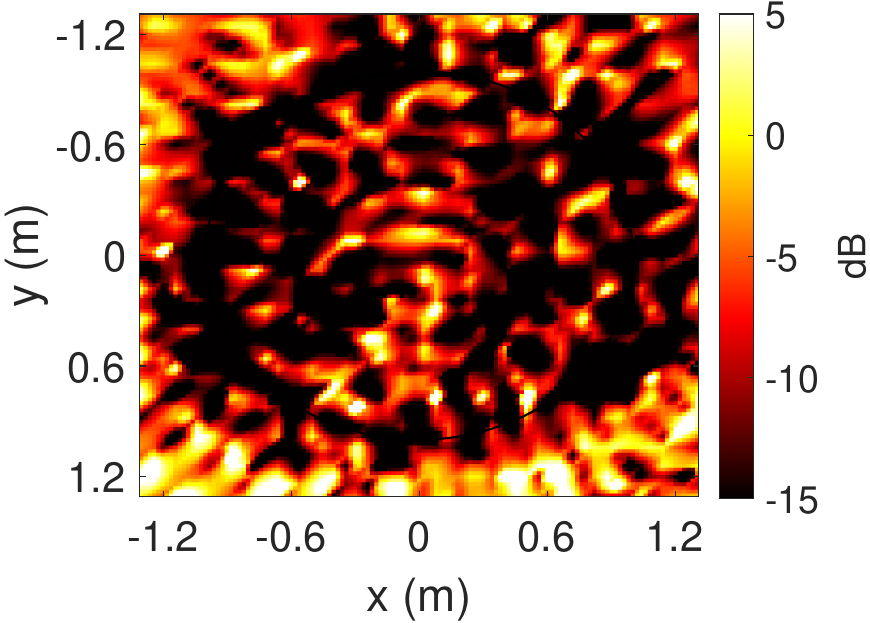}%
\label{figure5e}}
\subfloat[]{\includegraphics[width=0.25\textwidth]{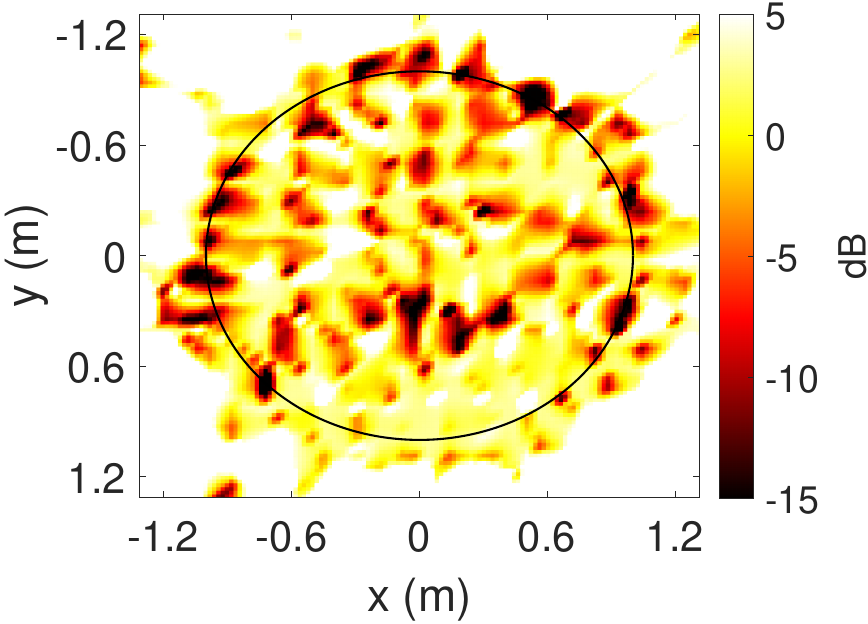}%
\label{figure5f}}
\subfloat[]{\includegraphics[width=0.25\textwidth]{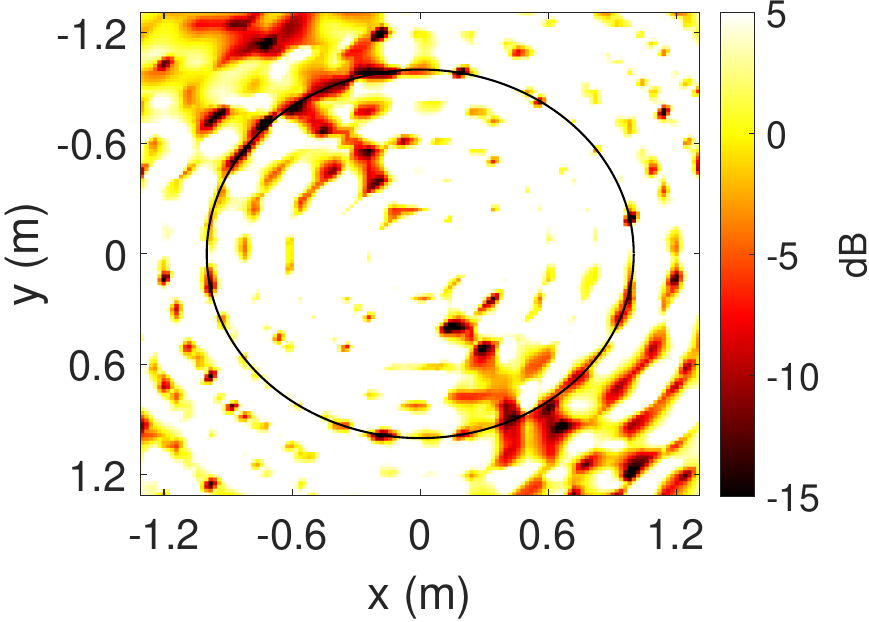}%
\label{figure5g}}
\caption{Sound field reconstruction and NSE distribution with circular microphone
distribution at 900 Hz for different methods. The target region is bounded by the black circle: (a) Original (b) Proposed (c) PINN (d) Harmonics-based (e) NSE of proposed (f) NSE of PINN (g) NSE of harmonics-based.}
\label{RecontructionCir}
\end{figure*}

The sound field reconstruction results and NSE distributions for different methods at $900$ Hz are shown in Fig.~\ref{RecontructionCir}, indicating the proposed method can reconstruct the sound field beyond the target region smoothly, demonstrating the proposed method can be generalized to extrapolate from measurements.
%out-of-sample scenarios. 
Based on Figs.~\ref{figure5c} and~\ref{figure5f}, the PINN method can only reconstruct the sound field over the target region but has significant errors for reconstructing the sound field outside of $\Omega$. 
%{\bf This limitation is caused by the distribution range of microphone observations of PDE loss evaluation points. - Need to rephrase! }
The reconstruction range is restricted by the distribution range of microphone observations and PDE evaluation points, which is one of the common limitations of typical PINNs.
%In the current PINN training strategy, PDE loss is calculated by 437 uniformly distributed points in the target region, meaning the nature of sound propagation is only learned in $\Omega$. The PDE loss can be evaluated in a larger region than $\Omega$, but it can cause longer training time and has little benefit in the estimation over $\Omega$.
As shown in Figs.~\ref{figure5d} and~\ref{figure5g}, the harmonics-based method can not correctly estimate the sound field at 900 Hz. 
%{\bf Cooment - delete - Is there a Bessel Zero at 900 Hz for the microphone radius?}.  

\subsection{Random microphone placement}\label{subsection4-3}
\begin{figure*}[!ht]
\centering
\subfloat[]{\includegraphics[width=0.48\textwidth]{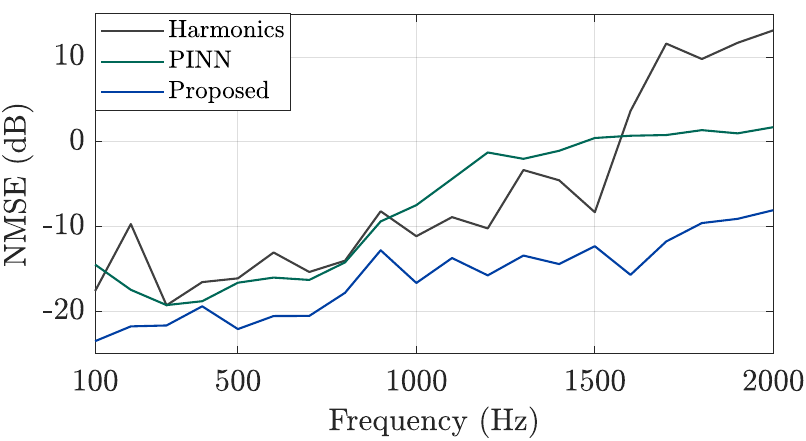}%
\label{figure6a}}
\hfill
\subfloat[]{\includegraphics[width=0.48\textwidth]{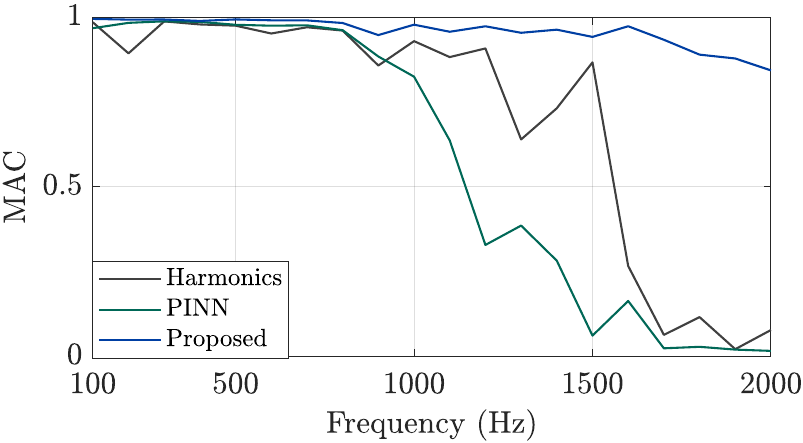}
\label{figure6b}}
\caption{NMSE and MAC with respect to the frequency with random microphone placement. (a) NMSE (b) MAC.}
\label{ErrorPlotRan}
\end{figure*}

In some cases, the circular microphone placement is not applicable. Therefore, we investigate the performance of different methods with the random microphone placement which can demonstrate the overall performance under more general cases. Results of NMSE and MAC evaluation are presented in Fig.~\ref{ErrorPlotRan}, illustrating the proposed method outperforms two competing methods for all frequencies, especially when the frequency is higher than 1500 Hz. Compared with Fig.~\ref{ErrorPlotCir}, we find that with random microphone placement i)the performance of the proposed method is improved by around 3 dB for frequencies under 500 Hz and reduced by around 4 dB for frequencies above 1700 Hz; ii) the outcome of PINN method is enhanced around 7 to 8 dB for frequencies below 1100 Hz, but for high frequencies, it cannot estimate the sound field correctly; iii)
the results of the harmonics-based method are more stable and improved by around 8 dB for frequencies below 1500 Hz, but the performance collapses after 1500 Hz.

Compared to circular microphone placement, the random microphone distribution samples across the whole target region can benefit the learning process of the proposed method and PINN method, especially for low frequencies. However, the random distribution results in a smaller number of samples in some areas, affecting the estimation of the complex sound field of high frequencies. For the harmonic-based method, the random microphone placement reduces the influence of Bessel zeros and enhance the estimation outcome. 

\begin{figure*}[!t]
\centering
\subfloat[]{\includegraphics[width=0.25\textwidth]{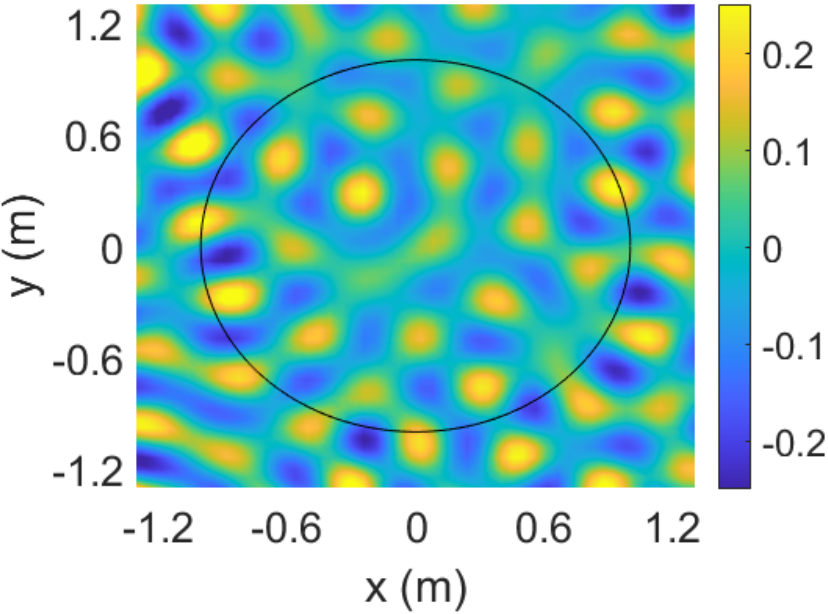}%
\label{figure7a}}
\subfloat[]{\includegraphics[width=0.25\textwidth]{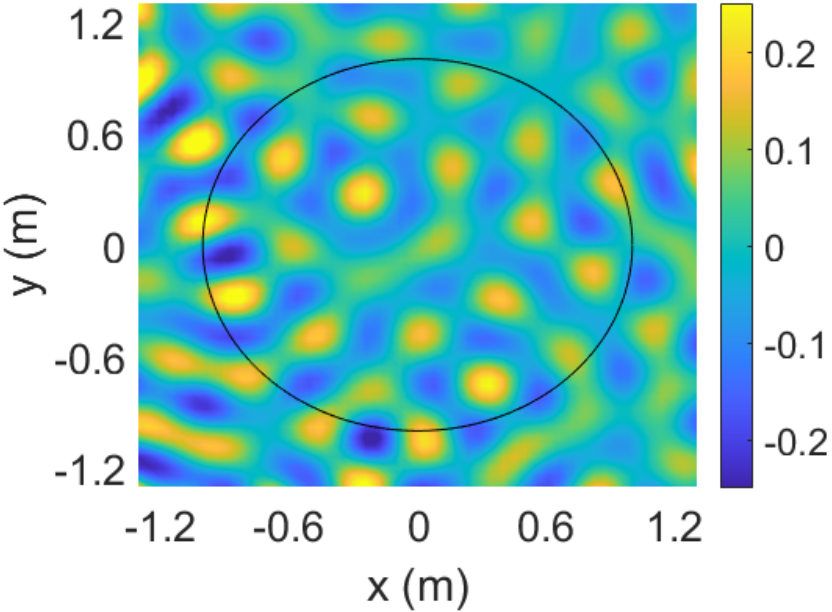}%
\label{figure7b}}
\subfloat[]{\includegraphics[width=0.25\textwidth]{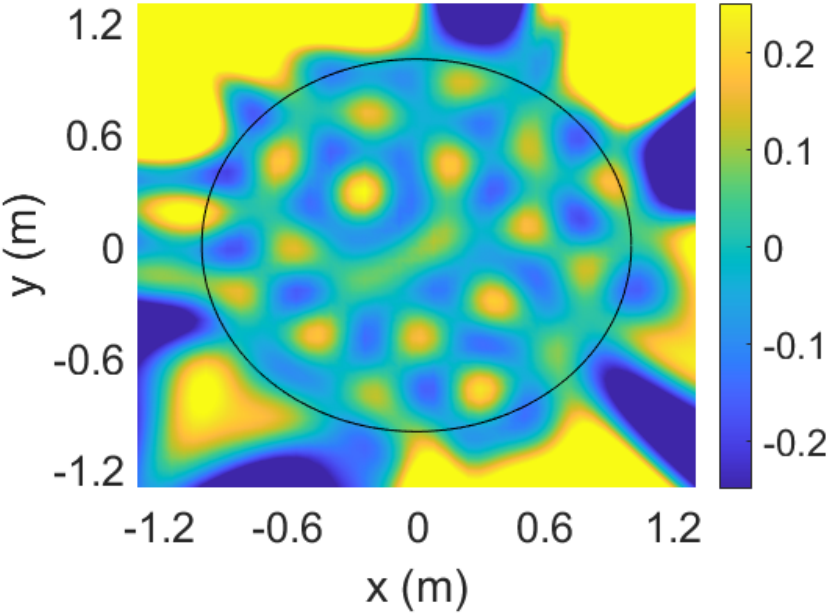}%
\label{figure7c}}
\subfloat[]{\includegraphics[width=0.25\textwidth]{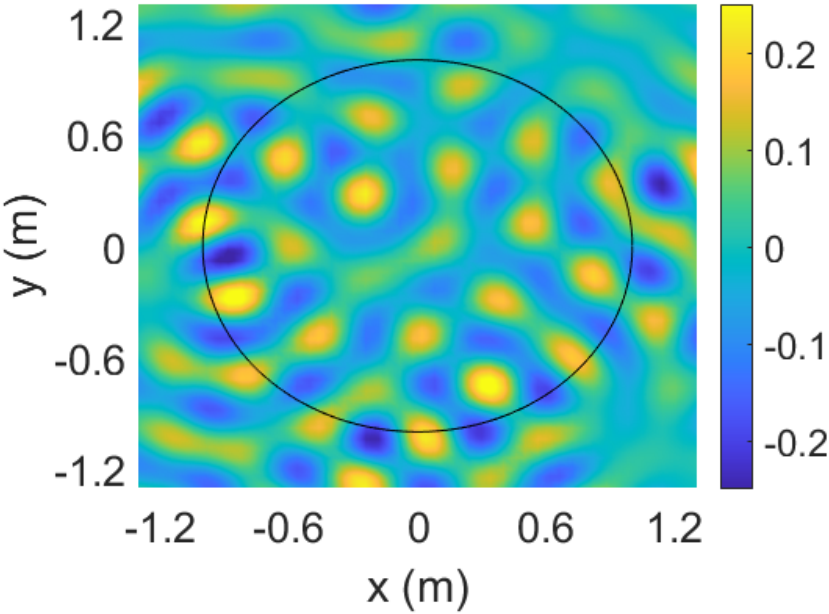}%
\label{figure7d}}
\hfill
\subfloat[]{\includegraphics[width=0.25\textwidth]{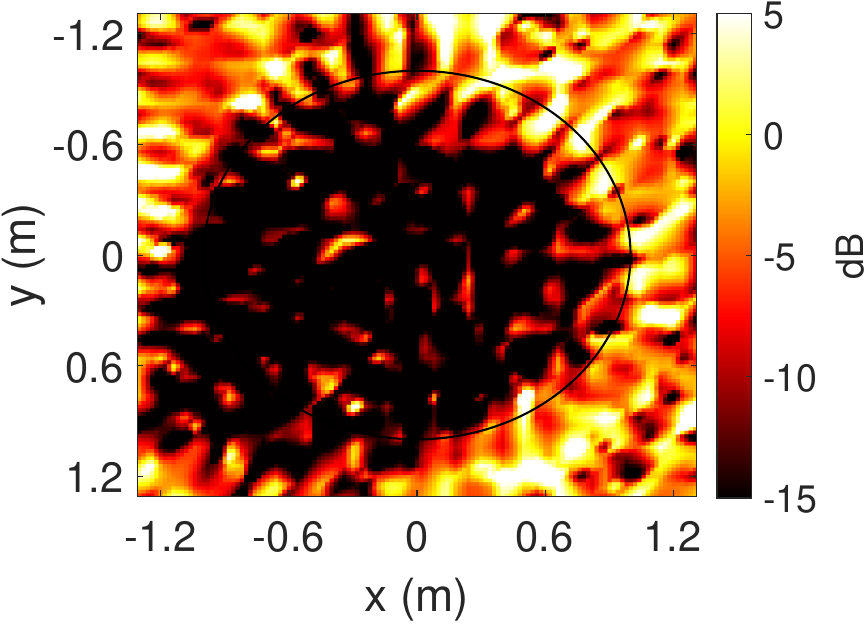}%
\label{figure7e}}
\subfloat[]{\includegraphics[width=0.25\textwidth]{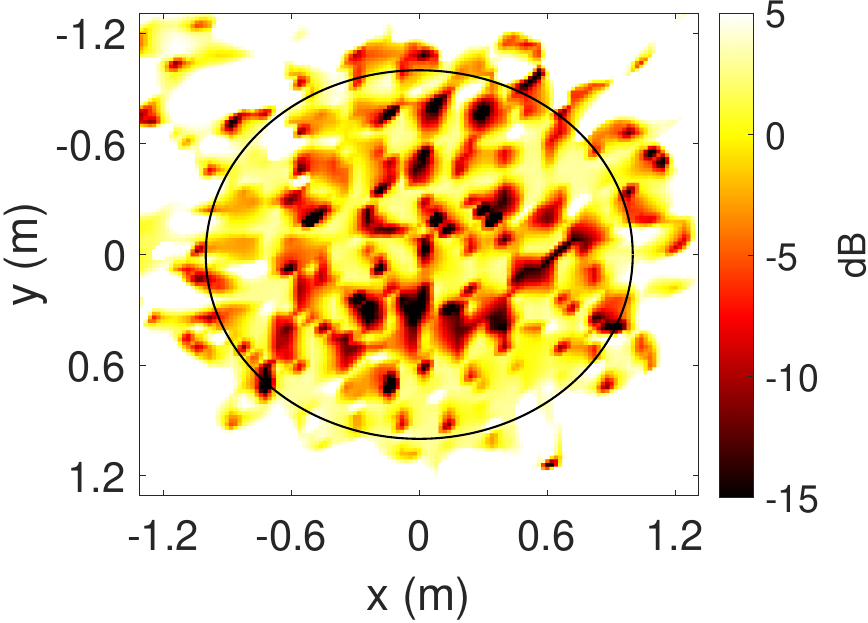}%
\label{figure7f}}
\subfloat[]{\includegraphics[width=0.25\textwidth]{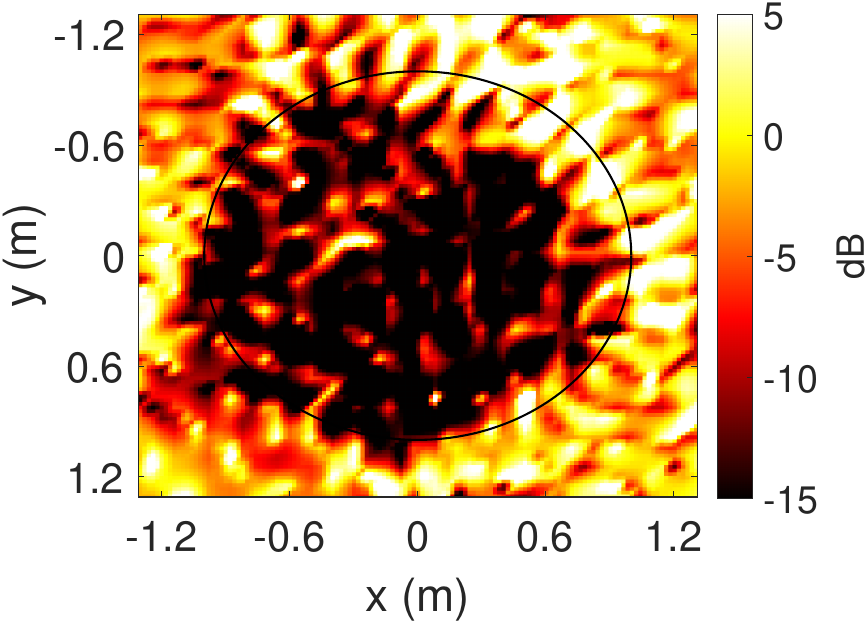}%
\label{figure7g}}

\caption{Sound field reconstruction and NSE distribution with random microphone distribution at 900 Hz for different methods. The target region is bounded by the black circle. (a) Original (b) Proposed (c) PINN (d) Harmonics-based (e) NSE of proposed (f) NSE of PINN (g) NSE of harmonics-based.}
\label{RecontructionRan}
\end{figure*}

 Similar to the investigation in Section \ref{subsection4-2}, we visualize the original sound field, reconstructed sound fields, and NSE distribution with random microphone placement at 900 Hz in Fig.~\ref{RecontructionRan}. Compared to Fig.~\ref{RecontructionCir}, we find that: i) the proposed method consistently has the best performance that reconstructs the sound field of the target region accurately; ii) the proposed method estimates the sound field larger than the target region; iii) compared to Figs.~\ref{figure5d} and~\ref{figure5g}, the reconstruction from harmonics based method is greatly improved with the random distribution and presents the accurate distribution larger than the target region; iv) the reconstruction from PINN method is slightly enhanced over the centre part of the target region.

 In section \ref{subsection4-2} and section \ref{subsection4-3}, we demonstrated that the proposed method outperforms two competing methods with both circular and random microphone placements. To further investigate the proposed method, we test its performance with more sparse microphone placement and different strengths of white Gaussian noise in the following subsections.

\subsection{Impact of varying the number of microphones} \label{sec:mic_num}
In this subsection, we investigate the performance of the proposed method under different microphone numbers with random placement. We keep all settings the same as in Section \ref{subsection4-3} except the number of microphones $Q$ which we vary for the following values: $Q= 35,~45,~55,~65, \mathrm{and}~75$.  
\begin{figure*}[!ht]
\centering
\subfloat[]{\includegraphics[width=0.48\textwidth]{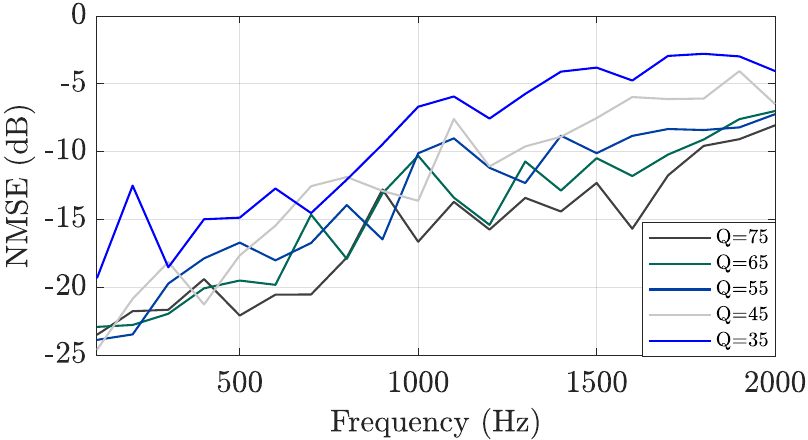}%
\label{figure8a}}
\hfill
\subfloat[]{\includegraphics[width=0.48\textwidth]{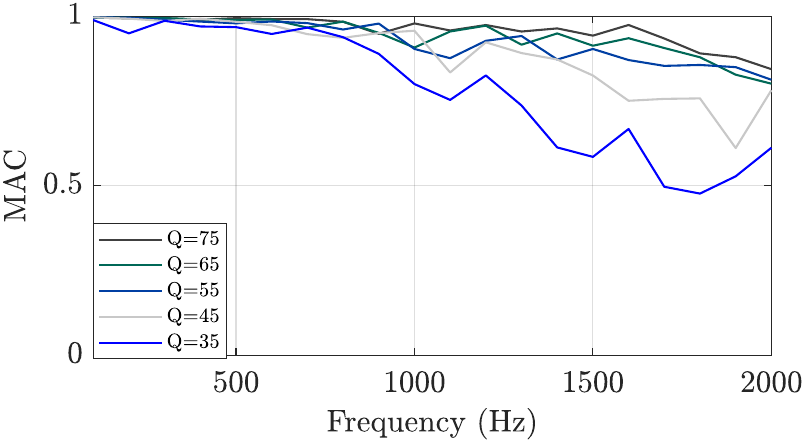}
\label{figure8b}}
\caption{(a) NMSE and (b) MAC with respect to the frequency with different numbers of randomly placed microphones. }
\label{ErrorPlotMic}
\end{figure*}

Results are shown in Fig.~\ref{ErrorPlotMic}.  Overall, the higher the number of microphones, the better the accuracy of the estimation of the soundield. 
%Overall, the estimation is more accurate with more number of microphones, and higher frequencies are more sensitive to the changes of $Q$. 
%The difference among $Q=55,~65, \mathrm{and}~75$ in both NMSE and MAC plots are very small, meaning with $Q\ge55$ the proposed method can accurately estimate the sound field for the whole range of interested frequencies. 

The difference between  NMSE and MAC values are very small for $Q=55,~65, \mathrm{and}~75$, which indicates for $Q\ge55$ the proposed method accurately estimates the sound field for the whole range of interested frequencies. 
For $Q=35$ and $45$, the proposed method accurately reconstructs the sound field under 1000 Hz and 1500 Hz respectively. The required microphone number is a function of the frequency and size of the target region, similar to the truncation rule in \cite{kennedy2007intrinsic,943347}, but further investigations are required to reveal the relationship among them. Compared to Fig.~\ref{ErrorPlotRan}, even with $Q=35$, the proposed method can still outperform two competing methods.

\subsection{Performance under different SNRs} \label{sec:different_snr}
We further investigate the robustness of the proposed method under different levels of white Gaussian noise. We keep all other settings the same as in Section \ref{subsection4-3} and only change the noise levels added to the microphone observations.  
\begin{figure*}[!ht]
\centering
\subfloat[]{\includegraphics[width=0.48\textwidth]{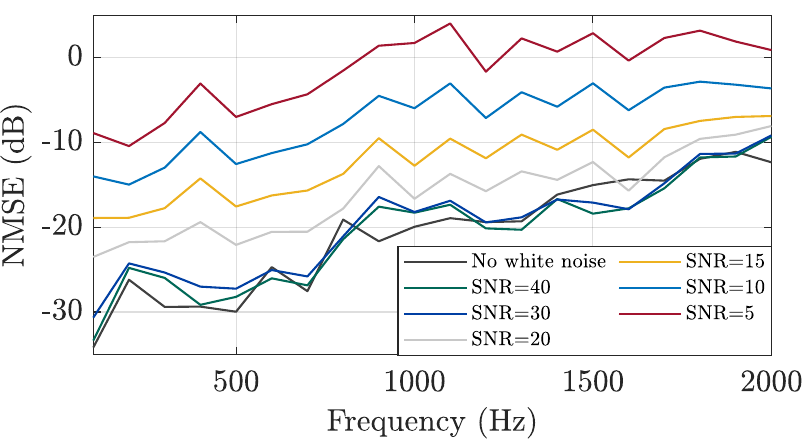}%
\label{figure9a}}
\hfill
\subfloat[]{\includegraphics[width=0.48\textwidth]{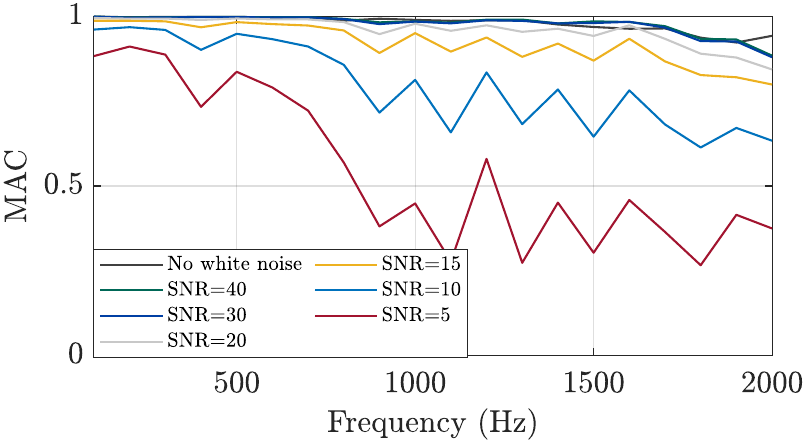}
\label{figure9b}}
\caption{(a) NMSE and (b) MAC with respect to the frequency with different noise levels under random microphone placements.}
\label{ErrorPlotSNR}
\end{figure*}

Results are presented in Fig.~\ref{ErrorPlotSNR}, where we find that: i) with noise under $30$ dB SNR, the proposed method has similar performance regardless of SNR, meaning a small level of white Gaussian noise has little influence on the proposed method; ii) when $\mathrm{SNR}\leq20~\mathrm{dB}$, the estimation accuracy decreases with the increment of noise level, especially for higher frequencies; iii) compared to the NMSE plot in Fig.~\ref{figure9a}, the MAC results, shown in Fig.~\ref{figure9b}, for high frequencies are more sensitive to noise, indicating the estimation for high frequencies tends to lose the overall similarity of the original sound field; iv) when $\mathrm{SNR}\leq20~\mathrm{dB}$, both NMSE and MAC plots among different noise level have similar shape, indicating that the outcome of the proposed method is consistent with frequencies; v) compared to Fig.~\ref{ErrorPlotRan}, even with $\mathrm{SNR}=15~\mathrm{dB}$, the proposed method still outperforms two competing methods.

\section{Conclusion}\label{sec5}

In this paper, we developed a new PINN architecture that embedded the fundamental solution of the wave equation into the network architecture, enabling the learned model to strictly satisfy the wave equation. The proposed point neuron learning method can estimate an arbitrary sound field based on microphone observations. Compared to other PINN methods, our approach can directly process complex numbers and is fully interpretable in physics, along with improved generalizability. The proposed method was evaluated in a sound field reconstruction problem within a reverberant environment with multiple evaluation metrics. Results indicate that the point neuron learning method outperformed two competing methods over all investigated frequencies. Additionally, the proposed method exhibited robustness to white Gaussian noise and proved effective even with sparse microphone observations. While we used sound field reconstruction problem only as an example, we expect the proposed point neuron architecture to be suitable for many applications involving wave propagation governed by the wave equation.

% \bmhead{Acknowledgements}

% Acknowledgements are not compulsory. Where included they should be brief. Grant or contribution numbers may be acknowledged.

% Please refer to Journal-level guidance for any specific requirements.

\section*{Declarations}

%Some journals require declarations to be submitted in a standardised format. Please check the Instructions for Authors of the journal to which you are submitting to see if you need to complete this section. If yes, your manuscript must contain the following sections under the heading `Declarations':

\begin{itemize}
\item Funding: This work is sponsored by the Australian Research Council (ARC) Discovery Projects funding schemes with project numbers DP200100693.
\item Competing interests: The authors declare that they have no competing interests.
\item Ethics approval and consent to participate: Not applicable.
\item Consent for publication: Not applicable.
\item Availability of data and materials: The data used and/or analysed during the current study are available from the corresponding author upon reasonable request.
\item Code availability: The code used and/or analysed during the current study is available from the corresponding author upon reasonable request.
\item Author contribution: TDA and HB formalized and conceptualized the problem. HB derived the formula and performed the experiments. TDA supervised the research. Both authors read and proved the published version of the manuscript.
\end{itemize}

\noindent
%If any of the sections are not relevant to your manuscript, please include the heading and write `Not applicable' for that section. 

%%===================================================%%
%% For presentation purpose, we have included        %%
%% \bigskip command. Please ignore this.             %%
%%===================================================%%
%\bigskip
%\begin{flushleft}%
%Editorial Policies for:

%\bigskip\noindent
%Springer journals and proceedings: \url{https://www.springer.com/gp/editorial-policies}

%\bigskip\noindent
%Nature Portfolio journals: \url{https://www.nature.com/nature-research/editorial-policies}

%\bigskip\noindent
%\textit{Scientific Reports}: \url{https://www.nature.com/srep/journal-policies/editorial-policies}

%\bigskip\noindent
%BMC journals: \url{https://www.biomedcentral.com/getpublished/editorial-policies}
%\end{flushleft}

\begin{appendices}
\section{Derivation of equation (\ref{gradient w})}\label{secA1}
We derive the back propagation of point neuron weights in this section. The frequency dependency $k$ and iteration index $n$ are omitted for notational simplicity.

With the input of $\boldsymbol x_{q}$, the output of the $v$-th point neuron can be expressed by
\begin{equation}\label{neuronV}
\mathcal{P}^{v}_{q}= w_{v}\frac{D_{v}}{D_{q}^{v}}e^{ik(D_{q}^{v}-D_{v})}.
\end{equation}
For updating point neuron weights, from (\ref{eq:2A}), we have
\begin{equation}\label{Gw0}
\frac{\partial \mathcal{L}}{\partial  w_{v}^{*}}=\frac{\partial \big(\sum^{Q}_{q=1}{| \mathcal{P}(\boldsymbol x_{q})-P(\boldsymbol x_{q}) |^{2}}\big)}{\partial  w_{v}^{*}}+\lambda\frac{\partial(\rVert \boldsymbol w \rVert_{1})}{\partial w_{v}^{*}}.
\end{equation}
The first term can be further expressed by \cite{sarason2007complex}
\begin{subequations}\label{Gw1}
\begin{align}
\frac{\partial \big(\sum^{Q}_{q=1}{ | \mathcal{P}(\boldsymbol x_{q})-P(\boldsymbol x_{q}) |^{2}}\big)}{\partial w_{v}^{*}}
=&\sum^{Q}_{q=1}\Big(\frac{\partial \big(({\mathcal{P}_{q}-P_{q}})({\mathcal{P}_{q}-P_{q}})^{*}\big)}{\partial\mathcal{P}_{q}}\frac{\partial\mathcal{P}_{q}}{\partial w_{v}^{*}} \nonumber\\
+&\frac{\partial \big(({\mathcal{P}_{q}-P_{q}})({\mathcal{P}_{q}-P_{q}})^{*}\big)}{\partial{\mathcal{P}_{q}}^*}\frac{\partial{\mathcal{P}_{q}}^*}{\partial w_{v}^{*}}\Big),
\end{align}
\end{subequations}
where $\mathcal{P}_{q}$ and ${P}_{q}$ represent $\mathcal{P}(\boldsymbol x_{q})$ and $P(\boldsymbol x_{q})$, respectively.
We simplify each item separately to get
\begin{subequations}\label{Gw2}
\begin{align}
\frac{\partial \big(({\mathcal{P}_{q}-P_{q}})({\mathcal{P}_{q}-P_{q}})^{*}\big)}{\partial{\mathcal{P}_{q}}}=&({\mathcal{P}_{q}-P_{q}})^{*},\\
\frac{\partial\mathcal{P}_{q}}{\partial w_{v}^{*}}=&0,\\
\frac{\partial \big(({\mathcal{P}_{q}-P_{q}})({\mathcal{P}_{q}-P_{q}})^{*}\big)}{\partial{\mathcal{P}_{q}}^*}=&({\mathcal{P}_{q}-P_{q}}),\\
\frac{\partial{\mathcal{P}_{q}}^*}{\partial w_{v}^{*}(n)}=&\sum^{V}_{v'=1}\frac{\partial{\mathcal{P}^{v'}_{q}}^*}{\partial w_{v}^{*}}=\frac{D_{v}}{D_{q}^{v}}e^{-ik(D_{q}^{v}-D_{v})}.
\end{align}
\end{subequations}
Substituting (\ref{Gw2}) into (\ref{Gw1})
\begin{equation}\label{Gw3}
\frac{\partial \big(\sum^{Q}_{q=1}{| \mathcal{P}(\boldsymbol x_{q})-P(\boldsymbol x_{q}) |^{2}} \big)}{\partial w_{v}^{*}}
=\sum^{Q}_{q=1}({\mathcal{P}_{q}-P_{q}})\frac{D_{v}}{D_{q}^{v}}e^{-ik(D_{q}^{v}-D_{v})}.
\end{equation}
For the model complexity loss, we have \cite{sarason2007complex}
\begin{subequations}\label{Gw4}
\begin{align}
\frac{\partial(\rVert \boldsymbol w \rVert_{1})}{\partial w_{v}^{*}}
=&\sum^{V}_{v'=1}\Big(\frac{\partial(|w_{v'}|)}{\partial w_{v}^{*}}\Big)=\frac{1}{2}\Big(\frac{\partial(|w_{v}|)}{\partial\operatorname{\mathbb{R}e}\{ w_{v}^{*}\}}+i\frac{\partial(|w_{v}|)}{\partial\operatorname{\mathbb{I}m}\{ w_{v}^{*}\}}\Big),
\end{align}
\end{subequations}
Given the absolute value of a complex number $|w_{v}|=\sqrt{\operatorname{\mathbb{R}e}\{ w_{v}^{*}\}^2+\operatorname{\mathbb{I}m}\{ w_{v}^{*}\}^2}$, where $\operatorname{\mathbb{R}e}\{ \cdot\}$ and $\operatorname{\mathbb{I}m}\{ \cdot\}$ denote the real and imaginary parts of the argument, we have
\begin{subequations}\label{Gw5}
\begin{align}
\frac{\partial(|w_{v}|)}{\partial\operatorname{\mathbb{R}e}\{ w_{v}^{*}\}}=&\frac{\operatorname{\mathbb{R}e}\{ w_{v}^{*}\}}{\sqrt{\operatorname{\mathbb{R}e}\{ w_{v}^{*}\}^2+\operatorname{\mathbb{I}m}\{ w_{v}^{*}\}^2}}=\mathrm{cos}\theta_{v},\\
\frac{\partial(|w_{v}|)}{\partial\operatorname{\mathbb{I}m}\{ w_{v}^{*}\}}=&\frac{\operatorname{\mathbb{I}m}\{ w_{v}^{*}\}}{\sqrt{\operatorname{\mathbb{R}e}\{ w_{v}^{*}\}^2+\operatorname{\mathbb{I}m}\{ w_{v}^{*}\}^2}}=\mathrm{sin}\theta_{v}.
\end{align}
\end{subequations}
Substituting (\ref{Gw5}) into (\ref{Gw4}) 
\begin{equation}\label{Gw6}
\frac{\partial(\rVert \boldsymbol w\rVert_{1})}{\partial w_{v}^{*}}=\frac{1}{2}e^{i\theta_{v}}.
\end{equation}
Based on (\ref{Gw0}), (\ref{Gw3}), and (\ref{Gw6}), we have 
\begin{equation}
\begin{aligned}
\label{Gw7}
\frac{\partial\mathcal{L}}{\partial w_{v}^{*}}=\sum^{Q}_{q=1}({\mathcal{P}_{q}-P_{q}})\frac{D_{v}}{D_{q}^{v}}e^{-ik(D_{q}^{v}-D_{v})}\,+\frac{1}{2}\lambda e^{i\theta_{v}}.
\end{aligned}
\end{equation}

\section{Derivation of equation (\ref{gradient Bx})}\label{secA2}
For updating biases, we take the derivation of $\partial \mathcal{L}(n)/\partial B_{v}^{x}(n)$ as an example, 
\begin{equation}\label{GB1}
\frac{\partial \mathcal{L}}{\partial B_{v}^{x}}=\frac{\partial(\sum^{Q}_{q=1}{| \mathcal{P}(\boldsymbol x_{q})-P(\boldsymbol x_{q}) |^{2}})}{\partial B_{v}^{x}}+\lambda\frac{\partial(\rVert \boldsymbol w \rVert_{1})}{\partial B_{v}^{x}}.
\end{equation}
As
\begin{equation}\label{GB2}
\frac{\partial(\rVert \boldsymbol w \rVert_{1})}{\partial B_{v}^{x}}=0,
\end{equation}
we have
\begin{subequations}\label{GB3}
\begin{align}
\frac{\partial \mathcal{L}}{\partial B_{v}^{x}}
=&\frac{\partial\big(\sum^{Q}_{q=1}{| \mathcal{P}(\boldsymbol x_{q})-P(\boldsymbol x_{q}) |^{2}}\big)}{\partial B_{v}^{x}}\\
=&\sum^{Q}_{q=1}\Big(\frac{\partial\big(({\mathcal{P}_{q}-P_{q}})({\mathcal{P}_{q}-P_{q}})^{*}\big)}{\partial\mathcal{P}_{q}}\frac{\partial\mathcal{P}_{q}}{\partial D_{q}^{v}}\frac{\partial D_{q}^{v}}{\partial B_{v}^{x}}\nonumber\\
+&\frac{\partial\big(({\mathcal{P}_{q}-P_{q}})({\mathcal{P}_{q}-P_{q}})^{*}\big)}{\partial\mathcal{P}_{q}}\frac{\partial\mathcal{P}_{q}}{\partial D_{v}}\frac{\partial D_{v}}{\partial B_{v}^{x}}\nonumber\\
+&\frac{\partial\big(({\mathcal{P}_{q}-P_{q}})({\mathcal{P}_{q}-P_{q}})^{*}\big)}{\partial\mathcal{P}_{q}^{*}}\frac{\partial\mathcal{P}_{q}^{*}}{\partial D_{q}^{v}}\frac{\partial D_{q}^{v}}{\partial B_{v}^{x}}\nonumber\\
+&\frac{\partial\big(({\mathcal{P}_{q}-P_{q}})({\mathcal{P}_{q}-P_{q}})^{*}\big)}{\partial\mathcal{P}_{q}^{*}}\frac{\partial\mathcal{P}_{q}^{*}}{\partial D_{v}}\frac{\partial D_{v}}{\partial B_{v}^{x}}\Big)
\end{align}
\end{subequations}
We calculate each item individually to get
\begin{subequations}\label{GB4}
\begin{align}
\frac{\partial\big(({\mathcal{P}_{q}-P_{q}})({\mathcal{P}_{q}-P_{q}})^{*}\big)}{\partial\mathcal{P}_{q}}=&({\mathcal{P}_{q}-P_{q}})^{*},\\
\frac{\partial\mathcal{P}_{q}}{\partial D_{q}^{v}}=&\sum^{V}_{v'=1}\frac{\partial\mathcal{P}_{q}^{v'}}{\partial D_{q}^{v}}\nonumber\\
=&w_{v}\frac{D_{v}}{{D_{q}^{v}}}e^{ik(D_{q}^{v}-D_{v})}\frac{(ikD_{q}^{v}-1)}{{D_{q}^{v}}},\\
\frac{\partial\mathcal{P}_{q}^{*}}{\partial D_{q}^{v}}=&\sum^{V}_{v'=1}\frac{\partial{\mathcal{P}_{q}^{v'}}^{*}}{\partial D_{q}^{v}}\nonumber\\
=&w_{v}^{*}\frac{D_{v}}{{D_{q}^{v}}}e^{-ik(D_{q}^{v}-D_{v})}\frac{(-ikD_{q}^{v}-1)}{{D_{q}^{v}}},\nonumber\\
=&(\frac{\partial\mathcal{P}_{q}}{\partial D_{q}^{v}})^{*},\\
\frac{\partial\mathcal{P}_{q}}{\partial D_{v}}=&\sum^{V}_{v'=1}\frac{\partial\mathcal{P}_{q}^{v'}}{\partial D_{v}}\nonumber\\
=&-w_{v}e^{ik(D_{q}^{v}-D_{v})}\frac{(ikD_{v}-1)}{D_{q}^{v}},\\
\frac{\partial\mathcal{P}_{q}^{*}}{\partial D_{v}}=&\sum^{V}_{v'=1}\frac{\partial{\mathcal{P}_{q}^{v'}}^{*}}{\partial D_{v}}\nonumber\\
=&{w_{v}}^{*}e^{-ik(D_{q}^{v}-D_{v})}\frac{(ikD_{v}+1)}{D_{q}^{v}},\nonumber\\
=&(\frac{\partial\mathcal{P}_{q}^{*}}{\partial D_{v}})^{*},\\
\frac{\partial D_{v}}{\partial B_{v}^{x}}=&\frac{B_{v}^{x}}{D_{v}},\\
\frac{\partial D_{q}^{v}}{\partial B_{v}^{x}}=&\frac{(B_{v}^{x}-x_{q})}{D_{q}^{v}}.
\end{align}
\end{subequations}
Substituting (\ref{GB4}) into (\ref{GB3}), we have
\begin{subequations}\label{GB5}
\begin{align}
\frac{\partial \mathcal{L}}{\partial B_{v}^{x}}
=&\sum_{q=1}^{Q}~2\operatorname{\mathbb{R}e}\Bigg\{({\mathcal{P}_{q}-P_{q}})^{*}w_{v}e^{ik(D_{v}^{q}-D_{v})}\frac{D_{v}}{D_{v}^{q}}\\
\times&\Big(\frac{-(ikD_{v}-1)}{{D_{v}}^{2}}B_{v}^{x}+\frac{(ikD_{v}^{q}-1)}{{D_{v}^{q}}^{2}}(B_{v}^{x}-x_{q})\Big)\Bigg\}.
\end{align}
\end{subequations}

The derivations of (\ref{gradient By}) and (\ref{gradient Bz}) are similar to the derivation of (\ref{gradient Bx}).
%% For submissions to Nature Portfolio Journals %%
%% please use the heading ``Extended Data''.   %%
%%=============================================%%

%%=============================================================%%
%% Sample for another appendix section			       %%
%%=============================================================%%

%% \section{Example of another appendix section}\label{secA2}%
%% Appendices may be used for helpful, supporting or essential material that would otherwise 
%% clutter, break up or be distracting to the text. Appendices can consist of sections, figures, 
%% tables and equations etc.

\end{appendices}

%%===========================================================================================%%
%% If you are submitting to one of the Nature Portfolio journals, using the eJP submission   %%
%% system, please include the references within the manuscript file itself. You may do this  %%
%% by copying the reference list from your .bbl file, paste it into the main manuscript .tex %%
%% file, and delete the associated \verb+\bibliography+ commands.                            %%
%%===========================================================================================%%

\bibliography{sn-bibliography}% common bib file
%% if required, the content of .bbl file can be included here once bbl is generated
%%\input sn-article.bbl

\end{document}